\documentclass[twoside]{article}

\usepackage[accepted]{aistats2022}

\usepackage[round]{natbib}
\renewcommand{\bibname}{References}
\renewcommand{\bibsection}{\subsubsection*{\bibname}}

\usepackage{url}
\usepackage{amssymb}
\usepackage{amsmath}
\usepackage{color,bbm,bm}
\usepackage{algorithmic}
\usepackage{algorithm}
\usepackage{graphicx}
\usepackage{enumitem}
\usepackage{booktabs}

\usepackage[usenames,dvipsnames]{xcolor}

\usepackage[colorlinks,linktoc=all]{hyperref}
\hypersetup{citecolor=Blue}
\hypersetup{linkcolor=Blue}
\hypersetup{urlcolor=Blue}
\usepackage{url}

\newtheorem{prop}{\protect\propositionname}

\usepackage{graphicx}

\providecommand{\definitionname}{Definition}
\providecommand{\assumptionname}{Assumption}
\providecommand{\lemmaname}{Lemma}
\providecommand{\propositionname}{Proposition}
\providecommand{\theoremname}{Theorem}

\newcommand{\scorex}
{s(x)}
\newcommand{\scorexk}
{s(x_k)}

\newcommand{\Exp}{\mathbb{E}}
\newcommand{\Var}{Var}
\newcommand{\Cov}{Cov}
\newcommand{\Real}[0]{\mathbb{R}}

\begin{document}

\runningtitle{Double Control Variates for Gradient Estimation in Discrete Latent Variable Models}

\twocolumn[

\aistatstitle{Double Control Variates for Gradient Estimation in \\
Discrete Latent Variable Models}

\aistatsauthor{ Michalis K. Titsias \And Jiaxin Shi }

\aistatsaddress{ DeepMind \And Microsoft Research New England} ]

\begin{abstract}
Stochastic gradient-based optimization for discrete latent variable models is challenging due to the high variance of gradients.   
We introduce a variance reduction technique for score function estimators that makes use of \emph{double control variates}. 
These control variates act on top of a main control variate, and try to further reduce 
the variance of the overall estimator. 
We develop a double control variate for the REINFORCE leave-one-out estimator using Taylor expansions. 
For training discrete latent variable models, such as variational autoencoders with binary latent variables, %
our approach adds no extra computational cost compared 
to standard training with the REINFORCE leave-one-out estimator. 
We apply our method to %
challenging high-dimensional toy examples 
and for training %
variational autoencoders with binary latent variables. 
We show that our estimator can have lower variance 
compared to other state-of-the-art estimators.
\end{abstract}

\section{INTRODUCTION}

Several problems in machine learning,  such as variational inference and reinforcement learning, 
require the optimization 
of an intractable expectation of an \emph{objective function} under a distribution with tunable parameters. 
Since exact gradients with respect to the parameters
of the distribution are intractable, 
optimization must rely on unbiased stochastic estimates. 
Pathwise or reparametrization gradients \citep{glasserman2003monte}
have been shown to be effective for machine learning problems \citep{Kingma2014,Rezende2014,Titsias2014_doubly}, but they are only applicable to 
continuous distributions.  
A more general class of gradient estimators based on the score function method or REINFORCE 
\citep{Glynn:1990, Williams92} is
applicable to both continuous and discrete distributions.
However, score function estimators suffer from high variance and reducing the variance remains an important open problem.  

Variance reduction techniques 
for REINFORCE estimators range from simple baselines \citep{Ranganath14,mnih2014neural} 
and Rao-blackwellization \citep{titsias2015local, tokui2017evaluating} 
to more advanced gradient-based  
control variates \citep{Tucker2017,Grathwohl2018,Guetal2016} and coupled sampling \citep{yin2018arm,disarm,yin2020,dimitrievzhou21}.
Another variance reduction method that 
has become prominent recently is the 
REINFORCE leave-one-out estimator
\citep{salimans2014using, Kool2019Buy4R, richter2020}, that assumes $K \geq 2$ samples and uses a leave-one-out procedure to define sample-specific stochastic control variates. Despite 
its simplicity, this estimator performs very strongly for training  
discrete latent variables models
\citep{disarm,richter2020,dongetal2021}. Presumably this is because the leave-one-out stochastic baselines can automatically adapt to the non-stationarity of the objective function which has trainable parameters itself, e.g., the %
parameters in the generative model.

In this work, our motivation is to take advantage of the compositional structure 
of control variate techniques \citep{owenbook, geffnerdomke18}, where multiple control variates can be linearly combined, to further reduce the variance of an existing estimator. Specifically,
we focus on the REINFORCE leave-one-out (RLOO) estimator and enhance it by adding extra control variates. %
We refer to the added baselines 
as \emph{double control variates} 
since they co-exist with the main RLOO 
baseline, and are designed to 
have a complementary effect by reducing the variance of the initial  RLOO estimator. 
We design the double control variates by applying Taylor expansions and  utilising gradients of the objective function over the Monte Carlo samples. For training latent variable models with discrete variables,
these gradients add essentially no extra  
computational cost since they can be obtained by the same backpropagation operation needed to collect the gradients over model parameters. 
Therefore, training by using our proposed estimator runs roughly at the same speed
with the previous RLOO approach. 

\begin{figure*}[!htbp]
\centering
\begin{tabular}{ccc}
{\includegraphics[scale=0.34]
{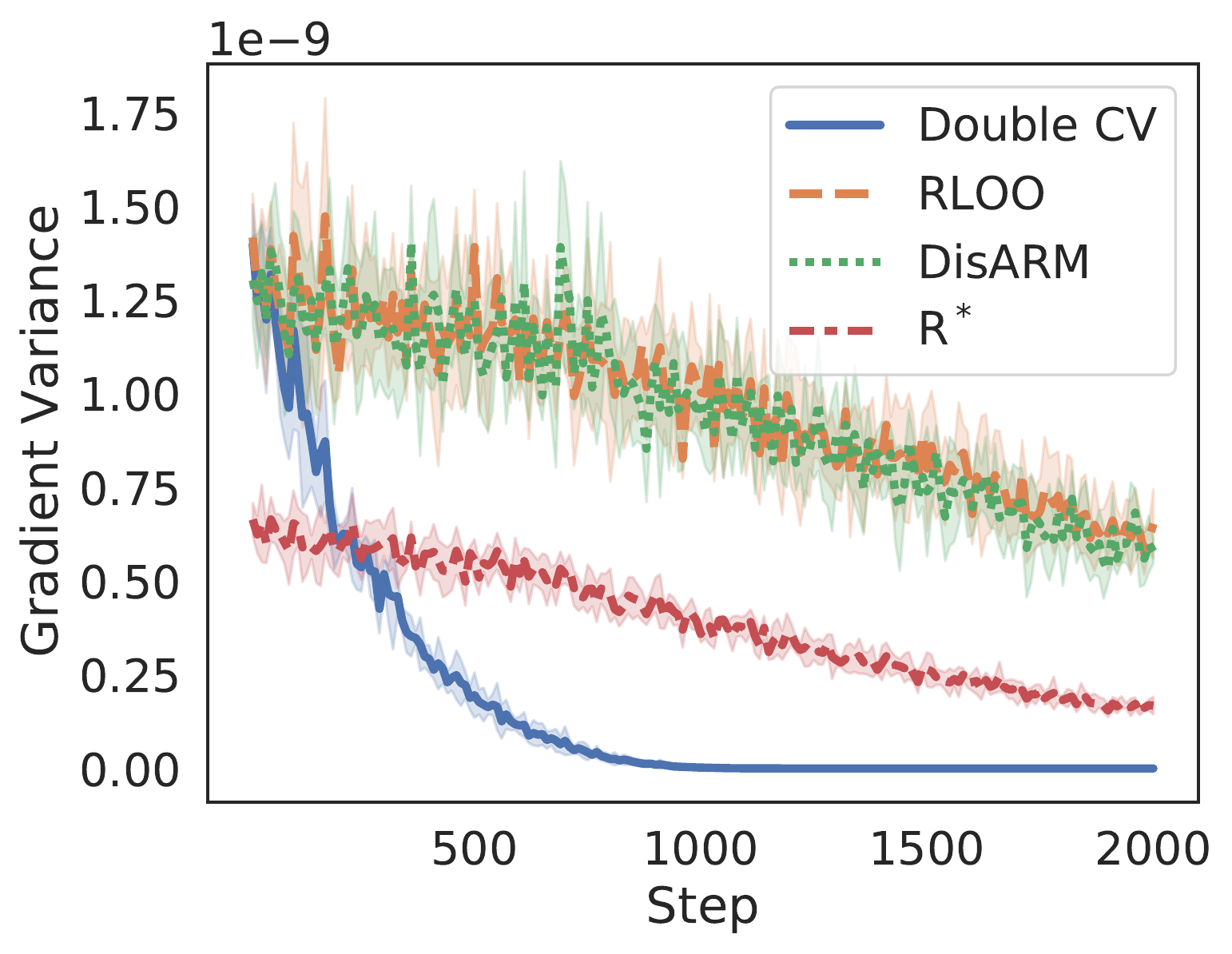}} &
{\includegraphics[scale=0.34]
{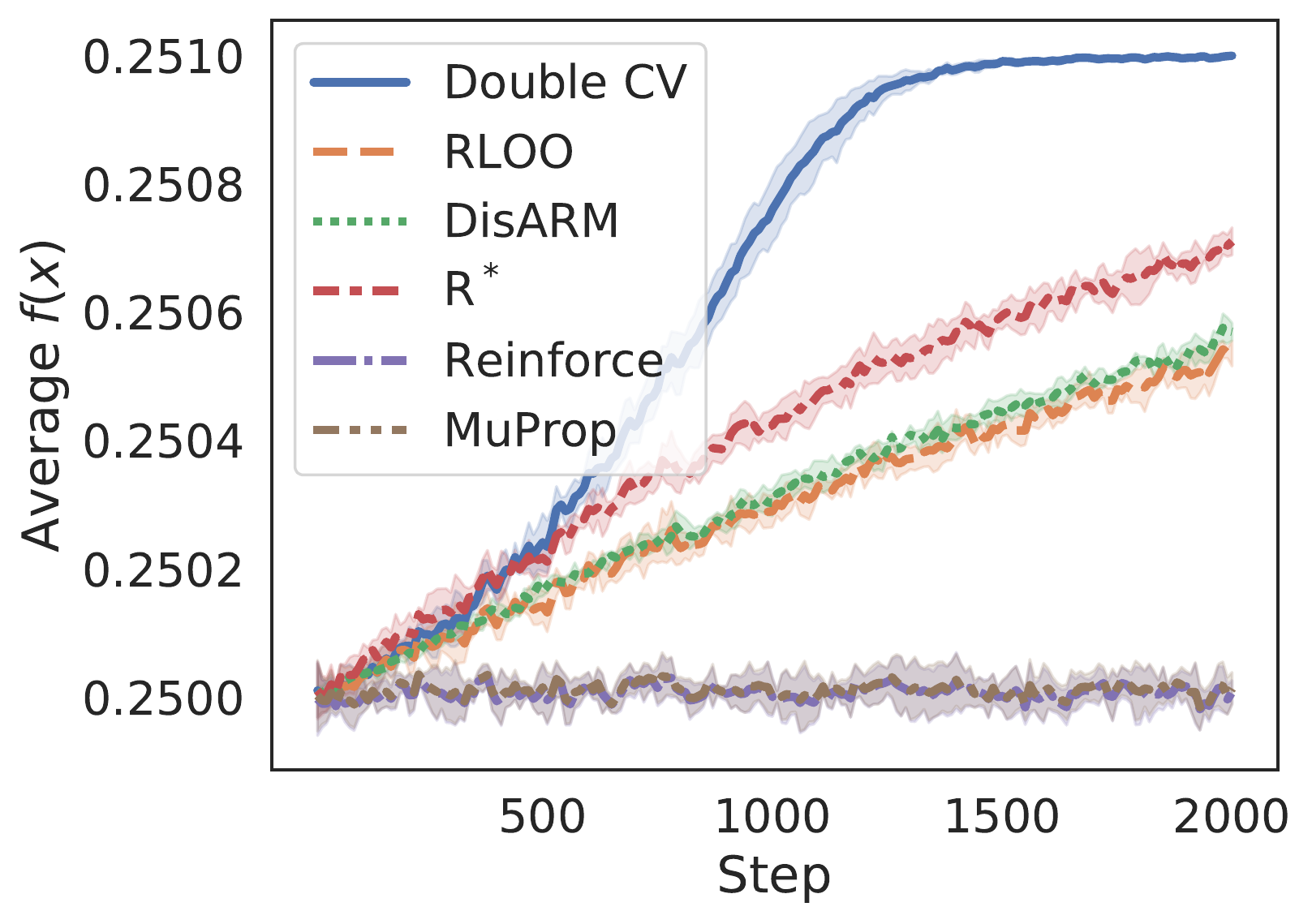}} 
&
{\includegraphics[scale=0.34]
{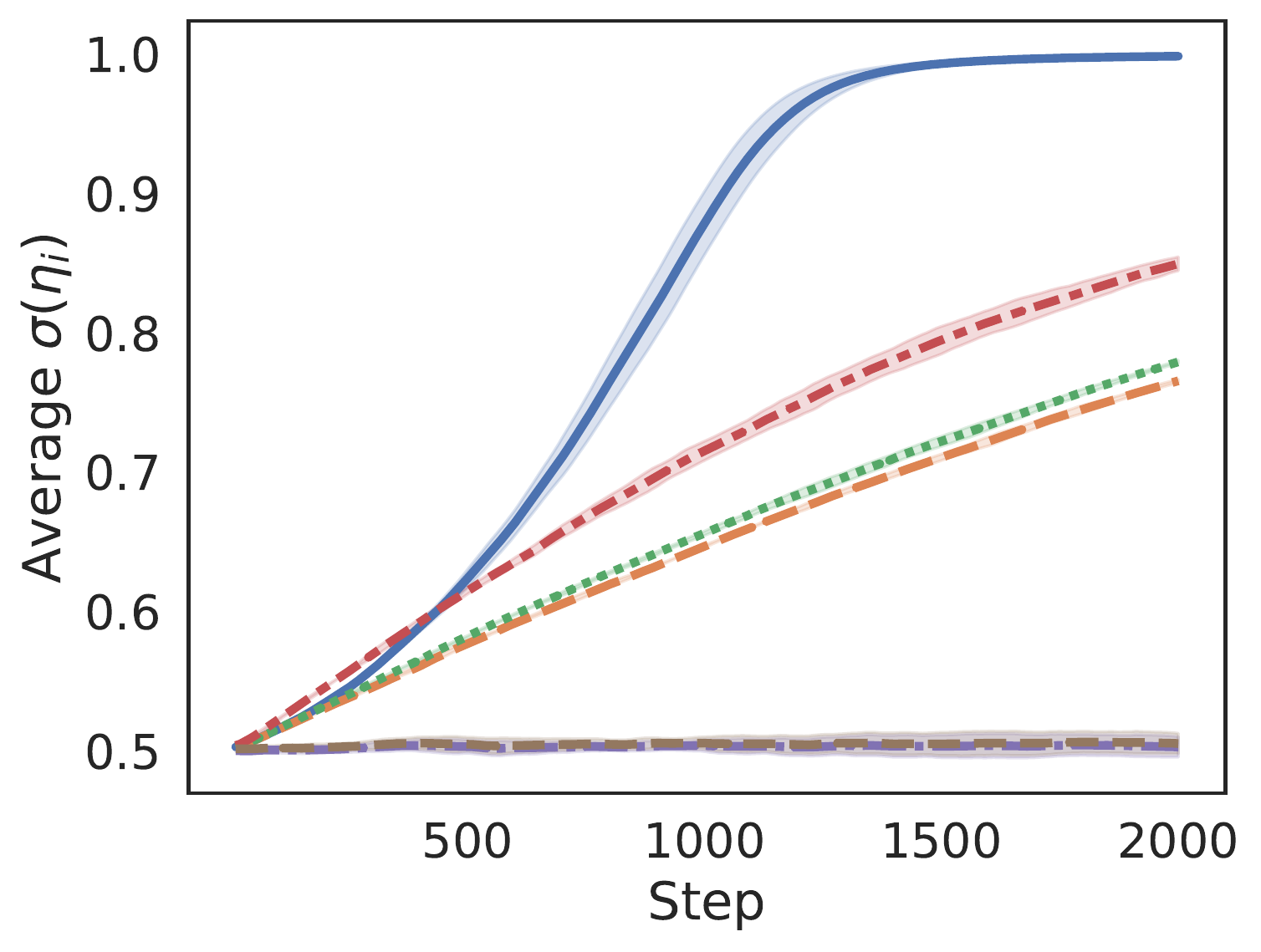}} 
\end{tabular}
\caption{
Variance reduction for a toy 200-dimensional maximization problem, following \cite{Tucker2017}, with binary variables and fitting probabilities $\sigma(\eta_i)$ (where $\sigma(\eta_i)=1$ is optimal); 
see Section \ref{sec:toyproblem}. 
\emph{Left:} Gradient variances %
for four different estimators. \emph{Middle:} Objective function that we want to maximize.  \emph{Right:} Average of the estimated  $\sigma(\eta_i)$s. In the latter two panels two additional estimators are shown.
The proposed double control variate estimator (Double CV) is the most effective one. 
} 
\label{fig:toy_example}
\end{figure*}   

We apply our double control variate approach to toy learning examples (see Fig.\ \ref{fig:toy_example}) and for training  
variational autoencoders with binary latent variables. 
We show that our estimator outperforms other methods including standard RLOO, DisARM~\citep{disarm,yin2020} and its improved version ARMS~\citep{dimitrievzhou21} when using $K = 2$ or more samples. 
Although we focus on binary latent variables in our experiments, our estimator is equally applicable to categorical latent variables.

\section{BACKRGOUND} 

Assume $f(x)$ is a differentiable objective function, where $x$ is a $D$-dimensional vector. 
We want to maximize the 
expectation 
$
\Exp_{q_\eta(x)} 
\left[ f(x) \right]
$
with respect to the parameters $\eta$ of
some distribution $q_\eta (x)$.  
Since $f(x)$ can have a complex non-linear form, the expectation is generally intractable. 
For instance, such problems arise in variational inference~\citep{blei2017variational}, where 
$f(x)$ is the instantaneous ELBO
and $q_\eta(x)$ the variational 
distribution, and in reinforcement learning, where $f(x)$ is a reward function and $q_\eta(x)$ is the policy~\citep{weaver2001optimal}.

To apply gradient-based optimization over $\eta$
we need to compute the gradient
\begin{equation}
\nabla_\eta \Exp_{q_\eta(x)} 
\left[ f(x) \right]
= \Exp_{q_\eta(x)} 
\left[ f(x) \nabla_\eta \log q_\eta(x)\right],
\label{eq:exactgradient}
\end{equation}
where for simplicity we assume $f(x)$ does not depend on  $\eta$.\footnote{If there is dependence this adds a low variance gradient to any stochastic estimator; see, e.g., \cite{disarm}.} Since this exact gradient is intractable, several techniques apply stochastic optimization
\citep{robbinsmonro51} based on unbiased Monte Carlo gradients by sampling from $q_\eta(x)$. A very general stochastic gradient is the score function or REINFORCE estimator  \citep{Glynn:1990,Williams92, Carbonetto2009, paisleyetal12, Ranganath14, mnih2014neural},
\begin{equation}
\frac{1}{K} \sum_{k=1}^K \left(f(x_k) - b \right)
\nabla_\eta \log q_\eta(x_k), \ \ x_k \sim q_\eta(x),
\end{equation}
where $b$ is called a \emph{baseline} and is often learned to reduce the variance. 
Given $K \geq 2$ samples, a powerful variant of this approach that avoids learning $b$ is the REINFORCE leave-one-out (RLOO) estimator \citep{salimans2014using,Kool2019Buy4R, richter2020} that takes advantage of multiple evaluations of $f$:
\begin{align}
\! \frac{1}{K} \! \sum_{k=1}^K \! \left(f(x_k) - \frac{1}{K-1}\sum_{
   j\neq k} f(x_j)\right)\nabla_\eta\log q_\eta(x_k), 
\label{eq:loogeneral}   
\end{align}
where each leave-one-out average 
$\frac{1}{K-1}\sum_{j\neq k} f(x_j)$
acts as a sample-specific control variate that excludes the current sample $x_k$, so that the whole estimator is unbiased. This estimator can also be re-written as an unbiased covariance estimator\footnote{Because of the score function property $\Exp_{q_\eta(x)} [\nabla \log q_\eta(x)]=0$ the exact gradient can be re-written as $\text{Cov}[f(x), \nabla_\eta \log q_\eta(x)] = \Exp_{q_\eta(x)} 
\left[ \left(f(x) - \Exp_{q_\eta(x)}[f(x)] \right) \nabla \log q_\eta(x)\right]$; see \cite{salimans2014using}.}, i.e.\ $\text{RLOO}(\eta) = \frac{1}{K-1}\sum_{k=1}^K \left(f(x_k) - \frac{1}{K}\sum_{
   j = 1}^K f(x_j)\right)\nabla_\eta\log q_\eta(x_k)$, which could be more convenient in implementation
   \citep{Kool2019Buy4R,richter2020}. 

RLOO was shown to have strong empirical performance,
 especially for discrete variable problems 
 \citep{disarm,Kool2019Buy4R,dongetal2021}. 
It has the attractive property 
that the sample-specific control variates
 automatically adapt to the non-stationarity of $f(x)$. 
Specifically, the function $f(x) := f_\theta(x)$ can often contain additional \emph{model parameters}
$\theta$  updated at each optimization step (for $\theta$ is straightforward to obtain low variance gradients), as for instance in variational autoencoders (VAEs) \citep{Kingma2014,Rezende2014}. 
Although $\theta$ is changing, the sample-specific control variate $\frac{1}{K-1}\sum_{j\neq k} f_{\theta}(x_j)$ always remains
an unbiased estimate of $\Exp_{q_\eta(x)}[f_{\theta}(x)]$. 

However, RLOO is still limited in how much variance reduction it can achieve, as stated in the following proposition which is proved in the Appendix.

\begin{prop}
Consider the estimator 
$
\text{R}^*(\eta) = \frac{1}{K}\sum_{k=1}^K \left(f(x_k) - \Exp f \right)\nabla_\eta\log q_\eta(x_k)$, where $\Exp f = \Exp_{q_\eta(x)}[f(x)]$ is a constant baseline across all samples. Then, 
$\Var(\text{RLOO}) \geq \Var(\text{R}^*)$.
\end{prop}

Thus, the performance of RLOO is bounded by $\text{R}^*$ which uses the mean $\Exp f$ (intractable in practice) as a constant baseline. However, an estimator with a constant baseline, even an ideal one as  $\text{R}^*$, can often have substantial variance in practice.  
For instance, in the toy example shown by Fig.~\ref{fig:toy_example}, where $\text{R}^*$ is tractable, we compare our proposed double control variates 
method with both RLOO and $\text{R}^*$, and show that our technique can  outperform
$\text{R}^*$ significantly. Our proposed estimator in Section \ref{sec:doublycontrols} tries to further reduce the variance 
of RLOO and  can be considered 
as using more general sample-specific baselines, 
as outlined in Section \ref{sec:more_general_sample_specific}. 

\subsection{More General Sample-Specific Control Variates
\label{sec:more_general_sample_specific}
}   

Let's denote  by $x_{1:K}$ all $K$ samples in the estimator. 
We say a baseline $\gamma_k(x_{1:K})$ is \emph{sample-specific} 
if it varies with the sample index $k$ in the Monte Carlo sum, i.e.\ $\gamma_k(x_{1:K})
\neq \gamma_j(x_{1:K})$ for $k \neq j$. 
Note that each $\gamma_k(x_{1:K})$ 
can depend on all samples including also the ``current" sample $x_k$. 
A general estimator with sample-specific control variates is written as 
\begin{align}
& \frac{1}{K}\sum_{k=1}^K
\left\{ \left(f(x_k) - \gamma_k(x_{1:K}) \right)\nabla_\eta\log q_\eta(x_k) \right\} \nonumber \\ 
& + \frac{1}{K} \sum_{k=1}^K  \Exp_{q_\eta(x_k)} 
[\gamma_k(x_{1:K}) \nabla_\eta \log q_\eta(x_k)].
\label{eq:general_sample_specific}
\end{align} 
The added sample-specific correction 
term $\Exp_{q_\eta(x_k)} 
[\gamma_k(x_{1:K}) \nabla_\eta \log q_\eta(x_k)]$ must be analytically tractable and  ensures that the gradient is unbiased. 
In some cases the correction term can be dropped, since it 
will have overall zero expectation, as stated next. 
\begin{prop}
 Let $x_{1:k-1,k+1:K}$ denote all samples excluding  $x_k$. If 
 $\Exp_{q_\eta(x_{1:k-1,k+1:K})}[\gamma_k(x_{1:K})] = \text{const}$, then 
 $\Exp_{q_\eta(x_{1:K})} 
[\gamma_k(x_{1:K}) \nabla_\eta \log q_\eta(x_k)] = 0$. 
\end{prop}
See Appendix for the proof.
A special case of this arises in REINFORCE LOO where the baseline
$\gamma_k(x_{1:k-1,k+1:K})$ does not depend on the current sample $x_k$. However, 
more effective estimators 
can have a $\gamma_k$ 
depending on the current sample $x_k$ as well. For instance, such an estimator is the second variant (see equation \eqref{eq:doublyEstimator3}) of the double control 
variates approach presented next.

\section{DOUBLE CONTROL VARIATES FOR REINFORCE LOO
\label{sec:doublycontrols}
}

In RLOO the sample-specific baseline $\frac{1}{K-1}\sum_{j\neq k} f(x_j)$ is constant
with respect to $x_k$. Also it 
is stochastic 
and as shown 
by Proposition 1 its variance is lower bounded by the estimator with $\Exp f$ as the baseline.
Therefore, there is scope to further reduce 
the variance of this estimator, and the approach we follow is to consider additional control variates.
We refer to these control 
variates 
as \emph{double} since they act on top of the main RLOO baseline. 
 We construct these new control variates along two directions:
 
 \begin{enumerate}[label=(\alph*)]
     \item We want to add a different type of control variate that depends 
on $x_k$ which may have a complementary effect to the main RLOO baseline.

\item Since the main baseline $\frac{1}{K-1}\sum_{
   j\neq k} f(x_j)$ is stochastic and thus has  
   variance, we can try to reduce the variance by adding a separate control variate  for each stochastic random term $f(x_j)$.
 \end{enumerate}
 In the remaining of Section \ref{sec:doublycontrols} we simplify notation by 
 using $\scorex := \nabla_\eta \log q_\eta(x)$ to denote the score function. 
 To accomplish 
 both (a) and (b) simultaneously we start 
 with the unbiased estimator 
\begin{equation}
    \frac{1}{K}  \sum_{k=1}^K  \left[ f(x_k) + \alpha b(x_k) \right] 
     \scorexk  - \alpha \Exp_{q_\eta(x)} [ b(x) \scorex],
\label{eq:doublecv_initial}     
\end{equation}
where we introduced 
a control variate $b(x_k)$, that depends on the current sample  $x_k$ 
and has analytic global correction 
$\Exp_{q_\eta(x)} [ b(x) \scorex]$. 
Then, to create a double control variate estimator we
treat $f(x) + \alpha b(x)$ 
as the ``new effective objective function''
and apply the leave-one-out procedure to it. This leads to the following unbiased  estimator 
\begin{align}
& \frac{1}{K} \! \! \sum_{k=1}^K \! \! \left[f(x_k) \! + \! \alpha \textcolor{blue}{b(x_k)} \! - \!  \frac{1}{K \! \! - \! \! 1}\sum_{j\neq k} \! (f(x_j)  \! + \! \alpha \textcolor{red}{b(x_j)} ) \! \right] \! \!
     \scorexk \nonumber \\
     & -  \alpha \Exp_{q_\eta(x)} [b(x) \scorex].
\label{eq:doublyEstimator1} 
\end{align}
The scalar $\alpha$ is a regression coefficient that can be further optimized to reduce the variance; see Section \ref{sec:furtherdetails}.
In the above estimator we have highlighted with blue the first 
appearance \textcolor{blue}{$b(x_k)$}, that can be thought of as a baseline paired with the value $f(x_k)$, and with red the second appearances \textcolor{red}{$b(x_j)$} paired with the remaining values $f(x_j)$ of the main RLOO baseline. Intuitively, 
$b(x_k)$ can be considered as targeting to reduce 
the variance of $f(x_k)$ and 
$b(x_j)$ the variance of $f(x_j)$. 

 In Sections \ref{sec:mean_field} and \ref{sec:no_extra_grads} we describe two ways to specify $b(x)$. For training latent variable models, such as VAEs, the second one will be the most practical since it adds no extra cost. The first method helps to introduce the idea and it is based on a mean field argument.        

\subsection{Mean Field Approach
\label{sec:mean_field}
} 

The optimal choice of $b(x)$  is to become an exact surrogate of $f(x)$.\footnote{In the estimator \eqref{eq:doublyEstimator1} this leads to zero variance when $\alpha=-1$.} This motivates to construct $b(x)$ by applying some tractable approximation to $f(x)$.
While any surrogate of $f(x)$ with a tractable global correction could work, next we focus on the case when $f(x)$ is differentiable w.r.t.\ the input $x$. Specifically, we assume the target function $f$ is implemented as differentiable function of real-valued inputs, but is restricted on a discrete subset of its domain. 
Then, we construct $b(x)$ from a first order Taylor approximation around the mean $\mu = \Exp_{q_\eta(x)}[x]$, so that 
$f(x) \approx f(\mu) + \nabla f(\mu)^\top (x - \mu) = b(x)$. Furthermore, 
observe that any constant term in $b(x)$ can be ignored because it cancels out in \eqref{eq:doublyEstimator1}. Thus, the constant $f(\mu)$ in the Taylor approximation can be dropped, yielding the double control variate
\begin{equation}
b(x) = \nabla f(\mu)^\top (x - \mu).
\label{eq:doubly_controls_mu}
\end{equation}
By substituting this in \eqref{eq:doublyEstimator1} we obtain the estimator 
\begin{align}
& \frac{1}{K} \sum_{k=1}^K \bigg[
f(x_k) + \alpha \nabla f(\mu)^\top (x_k - \mu)
\nonumber \\
& -   \frac{1}{K \! - \! 1}\sum_{j\neq k}\left(f(x_j)  + \alpha \nabla f(\mu)^\top (x_j - \mu) \right) \! \bigg]
     s(x_k) \nonumber \\
    & - \alpha  \Exp_{q_\eta(x)}[ s(x) (x-\mu)^\top]
    \nabla f(\mu),
\label{eq:doublyEstimator2} 
\end{align}
where $\Exp_{q_\eta(x)}[s(x) (x-\mu)^\top]$ will typically be analytically tractable. 
For instance, for binary latent variables $x \in \{0,1\}^d$ and a factorised 
Bernoulli distribution 
\begin{equation}
q_\eta(x) =
\prod_{i=1}^d \mu_i^{x_i} 
(1-\mu_i)^{1-x_i}, \ \ \mu_i =\sigma(\eta_i),
\end{equation}
$\Exp_{q_\eta(x)}[s(x) (x -\mu)^\top] = \text{diag}(\mu \circ (1-\mu))$ and the global correction term simplifies to
$- \alpha  \mu \circ (1 - \mu) \circ \nabla f(\mu)$, where $\circ$ 
denotes element-wise vector product.

\subsection{An Estimator without Extra Gradient Evaluations
\label{sec:no_extra_grads}
}

The estimator in Eq.\ 
\eqref{eq:doublyEstimator2} requires a backpropagation operation to compute the gradient $\nabla f(\mu)$, which adds extra computational cost compared to standard RLOO. Next, we wish to develop an alternative 
estimator that avoids this extra cost for certain problems. For many applications, such as VAEs, the function $f(x)$ depends on model parameters 
$\theta$ (typically different than $\eta$) that we update 
at each optimization iteration by computing the gradients $\{\nabla_\theta f(x_j)\}_{j=1}^K$. Then, from the same backpropagation operations it is easy to also return the gradients with respect to the latent vectors, i.e. to compute 
$\{\nabla_{x_j} f(x_j) \}_{j=1}^K$. To simplify 
notation we will write $\nabla f(x) := \nabla_x f(x)$.
We would like to utilize these latter gradients to define the double control variate $b(x)$.  

Starting from \eqref{eq:doubly_controls_mu} 
we first want to modify $b(x_k)$ by replacing $\nabla f(\mu)$ with some new gradient computed from $\{\nabla f(x_j) \}_{j=1}^K$. We cannot 
use the full average because this will lead to $( \frac{1}{K}\sum_{j = 1}^K \nabla f(x_j) )^\top (x_k - \mu)$ which has an intractable global correction due to the 
intractable term $\Exp_{q_\eta(x_k)} [\nabla f(x_k)^\top (x_k - \mu) 
\nabla_\eta \log q_\eta(x_k)]$.  
However, we can use the leave-one-out gradient, i.e.\ 
by leaving out $\nabla f(x_k)$, which gives 
\begin{equation}
b_k(x_{1:K}) = \left( \frac{1}{K-1}\sum_{j \neq k} \nabla f(x_j) \right)^\top (x_k - \mu),
\label{eq:bk}
\end{equation}
where we used the index $k$ in $b_k$ to emphasize that this now becomes a sample-specific control variate that varies with sample index; see Section \ref{sec:more_general_sample_specific}.
This has a tractable correction term
$\Exp_{q_\eta(x_k)} 
[b_k(x_{1:K}) s(x_k)] 
$ and also satisfies $\Exp_{q_\eta(x_k)}[b_k(x_{1:K}] = 0$. 
Having specified the double control variate we express the unbiased estimator as stated below. 
\begin{prop} For $b_k(x_{1:K})$ from \eqref{eq:bk}  we obtain the following unbiased gradient estimator 
\begin{align}
& \frac{1}{K} \! \! \sum_{k=1}^K \! \! \bigg[\! f(x_k) \! + \! \alpha b_k(x_{1:K})
\! - \! \frac{1}{K \! \! - \! \! 1} \! \sum_{j\neq k} \! \left(f(x_j) \! + \! \alpha b_j(x_{1:K}) \right) \! \! \bigg]
 \nonumber \\ 
& \! \times \! s(x_{k}) \!  - \! \alpha  \Exp_{q_\eta(x)}[s(x) (x\! - \! \mu)^\top \!] \! 
    \left(\! \frac{1}{K} \! \sum_{k=1}^K \! \nabla f(x_k) \!  \right).
\label{eq:doublyEstimator3} 
\end{align}
\end{prop}
The proof of unbiasedness is given in the Appendix. Notably, the above estimator follows the general structure from Eq.\ \eqref{eq:general_sample_specific} for a certain choice of the sample-specific control variate.

\subsection{Further Details and Algorithmic Summary
\label{sec:furtherdetails}
} 

To apply the estimator in \eqref{eq:doublyEstimator3} 
we need to specify the regression coefficient $\alpha$ by minimizing 
the variance. 
If $g(\eta; \alpha)$ denotes the stochastic gradient and $\bar{g} = \Exp[g(\eta; \alpha)]$ the exact gradient where the latter does not depend on $\alpha$, 
the total variance is $\text{Tr}[\Exp(g(\eta; \alpha)-\bar{g})(g(\eta; \alpha)-\bar{g})^\top] = \Exp[||g(\eta; \alpha)||^2] + const$. 
Thus, in practice at each optimization iteration we can perform a gradient step towards minimizing the empirical variance $||g(\eta; \alpha)||^2$. 
There also exists an analytic formula (but requiring intractable expectations) for the optimal value of $\alpha$ that can inspire different types of learning rules; see Appendix for further details. 
The whole algorithm that also deals with a non-stationary $f_\theta(x)$, i.e., that includes $\theta$ updates at each iteration,
is outlined in Algorithm \ref{alg:doublecv}.      
For the special case where $K=2$ %
the estimator 
\eqref{eq:doublyEstimator3} simplifies as 
\begin{align}
& \Delta(x_1,x_2,\alpha)  
\frac{ \nabla_\eta \log q_\eta(x_1) \!  - \! \nabla_\eta \log q_\eta(x_2)}{2} \nonumber \\
& - \alpha  \Exp_{q_\eta(x)}[s(x)  (x-\mu)^\top]
    \frac{\nabla f(x_1) + \nabla f(x_2)}{2},
\label{eq:doublyEstimatorK=2} 
\end{align}
where $\Delta(x_1,x_2,\alpha) = f(x_1) - f(x_2) + \alpha [\nabla f(x_2)^\top (x_1 -\mu) - \nabla f(x_1)^\top (x_2 -\mu)]$. In the experiments we compare this estimator with
the %
DisARM method \citep{disarm,yin2020} that uses $K=2$ antithetic samples, and also with RLOO with $K=2$ samples.  

\begin{algorithm}[tb]
   \caption{Optimization with double control variate gradients}
   \label{alg:doublecv}
\begin{algorithmic}
\STATE {\bfseries input:}
   loss $f_\theta(x)$, distribution $q_\eta(x)$.
\STATE  Initialise $\theta$, $\eta$, $\alpha=0$. \\
\STATE {\bf for} $t=1,2,3,\ldots,$ {\bf do}
\end{algorithmic}
   \begin{algorithmic}[1]
   \STATE Draw $K$ samples $x_{1:K}$,  $x_k \sim q_\eta (x)$.
   \STATE Compute $\mu = \Exp_{q_{\eta}(x)}[x]$.
   \STATE %
   $[f(x_k), \! \! \nabla_\theta f(x_k), \! \! \nabla_{x_k} f(x_k)]_{k=1}^K \leftarrow  \! \text{grad}(f_\theta, q_\eta, x_{1:k}\!)$.
   \STATE Compute double control variates
   $b_k(x_{1:K})$ from Eq.\ \eqref{eq:bk}.
   \STATE Compute double control variates gradient $g(\eta; \alpha)$ from Eq.\ \eqref{eq:doublyEstimator3}.
   \STATE Adapt $\eta$: 
   $\eta \leftarrow \eta - \rho_t \times g(\eta; \alpha)$. 
   \STATE Adapt $\theta$: $\theta \leftarrow \theta - \hat{\rho}_t \times \frac{1}{K} \sum_{k=1}^K \nabla_{\theta} f_\theta(x_k)$. 
   \STATE Adapt regression scalar $\alpha$ by applying a gradient step to minimize $||g(\eta; \alpha)||^2$. 
   \end{algorithmic}
\begin{algorithmic}   
\STATE {\bf end for}  
\end{algorithmic}
\end{algorithm}

\section{RELATED WORK 
\label{sec:related}}

\begin{table*}[t]
\caption{Training nonlinear binary latent VAEs with $K=2$ (except RELAX which needs 3 evaluations of $f$) on MNIST, Fashion-MNIST, and Omniglot. %
We report the average training ELBO over 5 independent runs.}
\label{tab:K2}
\vskip 0.1in
\footnotesize
\setlength{\tabcolsep}{5pt}
\centering
\resizebox{\textwidth}{!}{%
\begin{tabular}{lrrrrrr}
\toprule
& \multicolumn{3}{c}{Bernoulli Likelihoods} & \multicolumn{3}{c}{Gaussian Likelihoods} \\
\cmidrule{2-4} \cmidrule(l){5-7}
& \multicolumn{1}{c}{MNIST} & \multicolumn{1}{c}{Fashion-MNIST}  & \multicolumn{1}{c}{Omniglot} & \multicolumn{1}{c}{MNIST} & \multicolumn{1}{c}{Fashion-MNIST}  & \multicolumn{1}{c}{Omniglot}  \\
\midrule
RLOO & $-103.11\pm0.16$ & $-241.53\pm0.24$ & $-116.83\pm0.05$ & $668.07\pm0.40$ & $179.52\pm0.23$ & $443.51\pm0.93$\\
Double CV & $\bf -102.45\pm0.13$ & $\bf -240.96\pm0.17$ & $\bf -116.22\pm0.08$ & $\bf 676.87\pm1.18$ & $\bf 186.35\pm0.64$ & $\bf 446.95\pm0.63$\\
DisARM & $-102.56\pm0.09$ & $-241.02\pm0.20$ & $-116.36\pm0.05$ & $668.03\pm0.61$ & $182.65\pm0.47$ & $446.22\pm1.38$\\
\midrule
RELAX (3 evals) & $\bf -101.86\pm0.11$ & $\bf -240.63\pm0.16$ & $\bf -115.79\pm0.06$ & $\bf 688.58\pm0.52$ & $\bf 196.38\pm0.66$ & $\bf 462.30\pm0.91$\\
\bottomrule
\end{tabular}
}
\end{table*}

\begin{figure*}[t]
\centering
\begin{tabular}{c}
{\includegraphics[width=0.96\textwidth]
{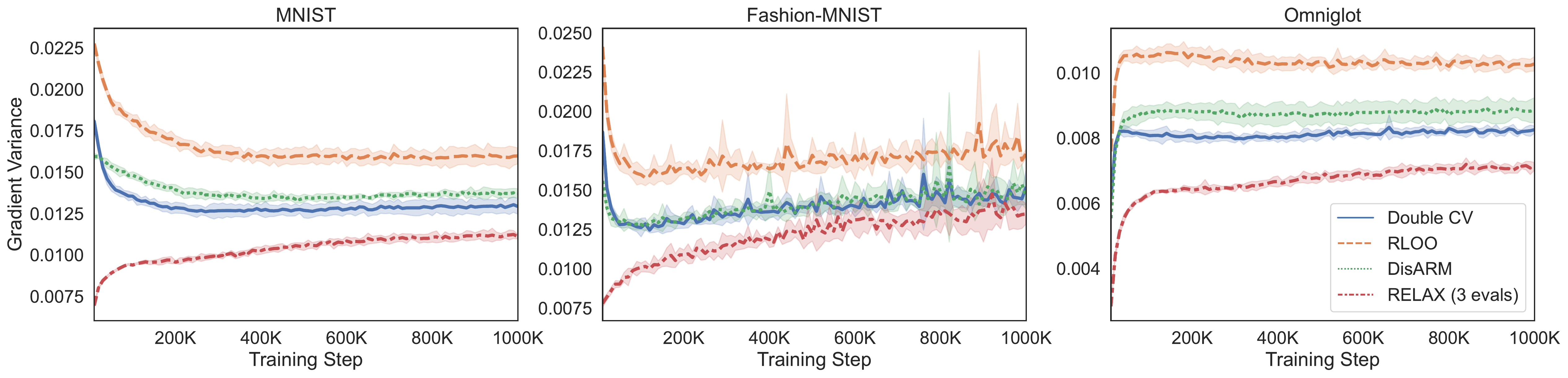}} \\
{\includegraphics[width=0.96\textwidth]
{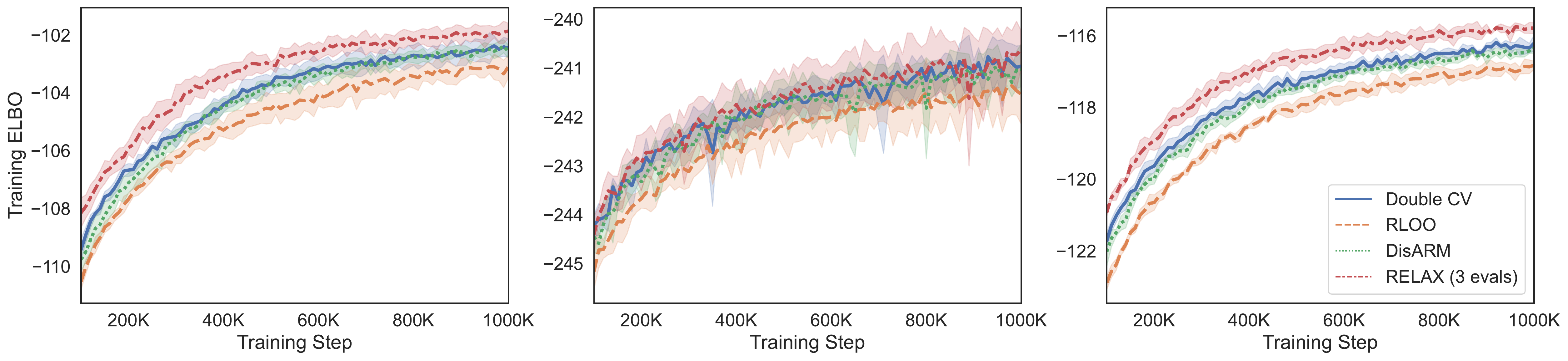}}
\end{tabular}
\vspace{-2.5mm}
\caption{
Training nonlinear binary latent VAEs with Bernoulli likelihoods with $K=2$ (except RELAX which needs 3 evaluations of $f$) on dynamically binarized MNIST, Fashion-MNIST, and Omniglot.
\emph{Top:} Variance of gradient estimates.
 \emph{Bottom:} Average ELBO on training examples.
} 
\label{fig:dyn-nonlinear-K2}
\end{figure*}

Our proposed gradient estimators 
follow the general form of unbiased REINFORCE 
estimators \citep{Williams92,Glynn:1990, Carbonetto2009, paisleyetal12,Ranganath14, mnih2014neural}, which unlike reparametrization or pathwise gradients \citep{Kingma2014,Rezende2014,Titsias2014_doubly}, are applicable also to discrete latent variables. The double control variates 
we develop build on top of the 
RLOO estimator
\citep{Kool2019Buy4R,salimans2014using,richter2020}; see also the VIMCO method of \cite{mnih2016variational} who also used a leave-one-out procedure. 
RLOO was shown to be a competitive estimator 
for challenging problems such as training 
VAEs with binary or categorical latent 
variables \citep{disarm,richter2020,dongetal2021}. As shown by our experiments, our enhancement of RLOO with double control variates leads to further variance reduction, and without increasing the computational cost when training VAEs. 

In our current framework, the double control variates are constructed by using the gradients of the objective function $f_\theta(x)$. 
These gradients are also used by other unbiased gradient techniques based on control variates, such as the MuProp estimator~\citep{Guetal2016}, %
REBAR~\citep{Tucker2017} and RELAX~\citep{Grathwohl2018}. 
Our method differs significantly since our control variates act on top of the sample-specific RLOO baseline $\frac{1}{K-1} \sum_{j \neq k} 
f_\theta(x_j)$, i.e., they try to have complementary effect to this existing control variate. 
This means that our estimators  preserve RLOO's property of capturing the non-stationarity of $f_\theta (x)$, since the leave-one-out baseline always tracks the expected value 
$\Exp [f_\theta(x)]$ as $\theta$ evolves. 
In contrast, previous gradient-based estimators use \emph{stand-alone} global control variates. 
For instance,  the baseline in MuProp \citep{Guetal2016} is constructed using only $f_\theta(\mu)$ and $x_k$, which can be a poor tracker of the expected value $\Exp [f_\theta(x)]$. Unlike MuProp, REBAR \citep{Tucker2017} is much 
more effective, however it is more expensive than our method --- it requires differentiating $f_\theta$ three times, while our method can work with just two, and it is less generally applicable since they assume a continuous reparameterization for $q$. 
RELAX~\citep{Grathwohl2018} suffers from the same problem as its strong performance relies on the REBAR control variate~\citep{richter2020}.

Other recent REINFORCE type of estimators for discrete latent variables are based on coupled
sampling \citep{owenbook}, such as antithetic sampling 
\citep{yin2018arm,disarm,yin2020, dimitrievzhou21}. 
For instance, the recent DisARM 
estimator independently proposed by 
\cite{disarm} and \cite{yin2020} 
was shown to give state-of-the-art results for binary latent-variable models with $K=2$ antithetic samples.

\section{EXPERIMENTS
\label{sec:experiments}
}

Code for reproducing all experiments is available at \url{https://github.com/thjashin/double-cv}. 

\begin{figure*}[!htbp]
\centering
\begin{tabular}{c}
{\includegraphics[width=0.95\textwidth]
{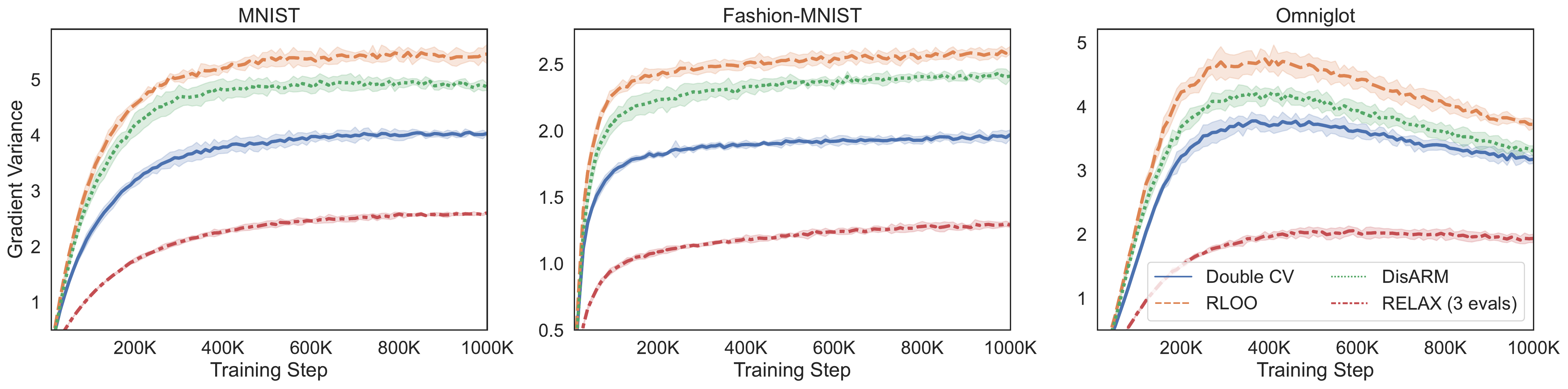}} \\
{\includegraphics[width=0.95\textwidth]
{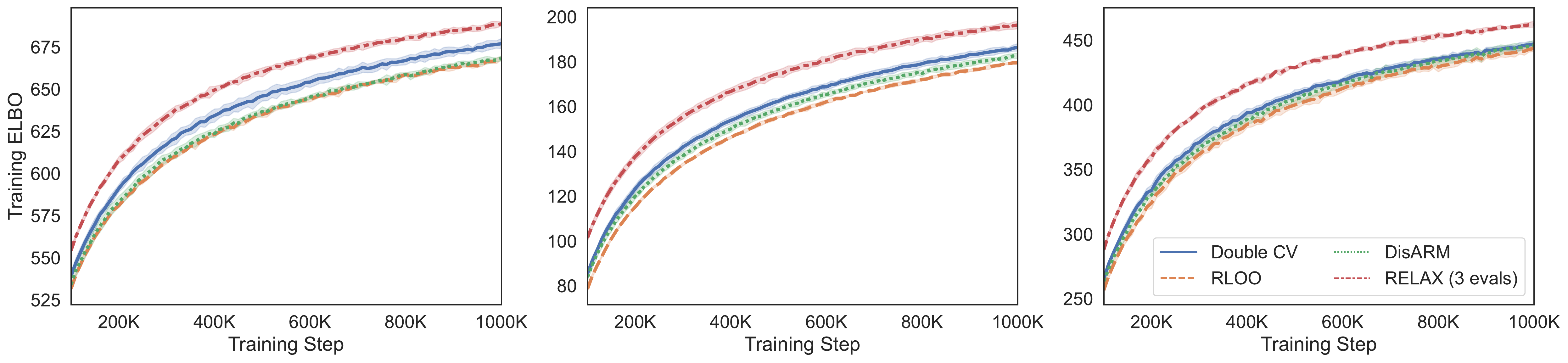}}
\end{tabular}
\caption{
Training nonlinear binary latent VAEs with Gaussian likelihoods with $K=2$ (except RELAX which needs 3 evaluations of $f$) on non-binarized MNIST, Fashion-MNIST, and Omniglot.
\emph{Top:} Variance of gradient estimates.
\emph{Bottom:} Average ELBO on training examples.
} 
\label{fig:cont-nonlinear-K2}
\end{figure*}

\subsection{Toy Learning Problem
\label{sec:toyproblem}
}

We consider a generalization 
of the artificial problem considered  by \cite{Tucker2017}. The goal is to maximize 
$
\mathcal{E}(\eta) = 
 \Exp_{q_\eta(x)}[D^{-1}\sum_{i=1}^D(x_i- p_0)^2], 
$
where $q_\eta (x) = \prod_{i=1}^D \sigma(\eta_i)^{x_i} (1 - \sigma(\eta_i))^{1-x_i}$, $p_0=0.499$
and the optimal solution is $\sigma(\eta_i)=1$ for all $i=1,\ldots,D$.  
While \cite{Tucker2017} considered $D=1$, here we additionally consider a more difficult high-dimensional case with  $D=200$. We compare five methods: 
 RLOO, DisARM,  MuProp, Reinforce (with no baselines)
and our proposed double control variates estimator (Double CV) from Eq.\ \eqref{eq:doublyEstimator3}. We use $K=2$ samples for all methods (note that Double CV in this case simplifies as in \eqref{eq:doublyEstimatorK=2}). 
Also we include in the comparison 
$\text{R}^*$ which is tractable in this toy example. 
Fig.~\ref{fig:toy_example}
compares the methods in terms 
of variance, the objective function
and the average value of the $D$ probabilities $\sigma(\eta_i)$.
Fig.~\ref{fig:toy_exampleD1} in the Appendix shows 
further comparison for the $D=1$ case,  
as in \cite{Tucker2017}. 
We observe that Double CV gradients have smaller variance which results in much faster optimization convergence. 

\subsection{Variational Autoencoders with Binary Latent Variables}
\label{sec:vae}

\paragraph{Experimental setup}
We consider training variational
autoencoders \citep{Kingma2014, Rezende2014} with binary latent variables. We conduct separate experiments for binary output data $y \in \{0,1\}^d$  
and continuous data $y \in \Real^d$.  
For binary data we 
use the standard Bernoulli likelihood. 
For continuous data we centered data between $[-1, 1]$ and consider a Gaussian 
likelihood of the form $p_\theta(y|x) = \mathcal{N}(y| m_\theta(x), \Sigma)$, where $m_\theta(x)$ is a decoder mean function that depends on the latent variable $x$ and $\Sigma$ is a learnable diagonal covariance matrix. We consider the datasets 
 MNIST, Fashion-MNIST and Omniglot. 
For all three datasets we use both the dynamically binarized versions and their original continuous versions.  

We consider 
the nonlinear VAE models used in \citet{yin2018arm,disarm}; results for linear VAEs are included in the Appendix. 
The VAE model uses fully connected neural networks with two hidden layers of $200$  LeakyReLU activation units
with the coefficient $0.3$. 
All models are trained using Adam
\citep{Adam} with learning rate 
$10^{-3}$ for the binarized data, 
while for the continuous data we used smaller learning rate 
$10^{-4}$. 
In all experiments the regression coefficient $\alpha$
of the double control variates was 
also trained (see Section \ref{sec:furtherdetails}) with Adam and with learning rate $10^{-3}$. 
For all 
experiments we use a uniform factorized 
Bernoulli prior over the $D=200$ dimensional latent variable $x$. 
The  model was trained by maximizing the ELBO using an amortised factorised variational Bernoulli distribution.

\begin{table*}[t]
\caption{Training a nonlinear binary latent VAE with $K=4$ (except RELAX which needs 3 evaluations of $f$) on MNIST, Fashion-MNIST, and Omniglot.
We report the average training ELBO over 5 independent runs.}
\label{tab:K4}
\vskip 0.1in
\footnotesize
\setlength{\tabcolsep}{5pt}
\centering
\resizebox{\textwidth}{!}{%
\begin{tabular}{lrrrrrr}
\toprule
& \multicolumn{3}{c}{Bernoulli Likelihoods} & \multicolumn{3}{c}{Gaussian Likelihoods} \\
\cmidrule{2-4} \cmidrule(l){5-7}
& \multicolumn{1}{c}{MNIST} & \multicolumn{1}{c}{Fashion-MNIST}  & \multicolumn{1}{c}{Omniglot} & \multicolumn{1}{c}{MNIST} & \multicolumn{1}{c}{Fashion-MNIST}  & \multicolumn{1}{c}{Omniglot}  \\
\midrule
RLOO & $-100.50\pm0.22$ & $-239.03\pm0.15$ & $-114.75\pm0.07$ & $687.83\pm0.50$ & $195.27\pm0.24$ & $460.23\pm1.42$\\
Double CV & $\bf -99.89\pm0.12$ & $-238.98\pm0.18$ & $\bf -114.56\pm0.06$ & $\bf 691.51\pm0.75$ & $\bf 199.01\pm0.60$ & $463.03\pm0.94$\\
DisARM & $-100.67\pm0.07$ & $-239.20\pm0.15$ & $-115.05\pm0.07$ & $683.28\pm0.89$ & $192.96\pm0.29$ & $458.38\pm0.88$\\
ARMS & $-100.07\pm0.08$ & $\bf -238.50\pm0.13$ & $-114.57\pm0.06$ & $687.26\pm1.21$ & $197.25\pm0.48$ & $\bf 463.30\pm0.86$\\
\midrule
RELAX (3 evals) & $-101.86\pm0.11$ & $-240.63\pm0.16$ & $-115.79\pm0.06$ & $688.58\pm0.52$ & $196.38\pm0.66$ & $462.30\pm0.91$\\
\bottomrule
\end{tabular}
}
\end{table*}

\begin{figure*}[h]
\centering
\begin{tabular}{c}
{\includegraphics[width=0.96\textwidth]
{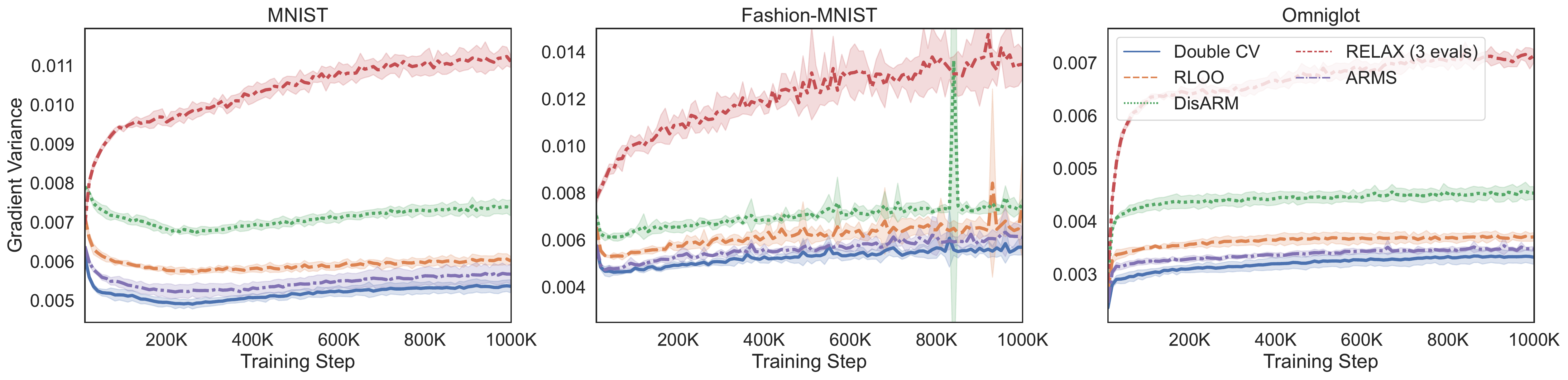}} \\
{\includegraphics[width=0.96\textwidth]
{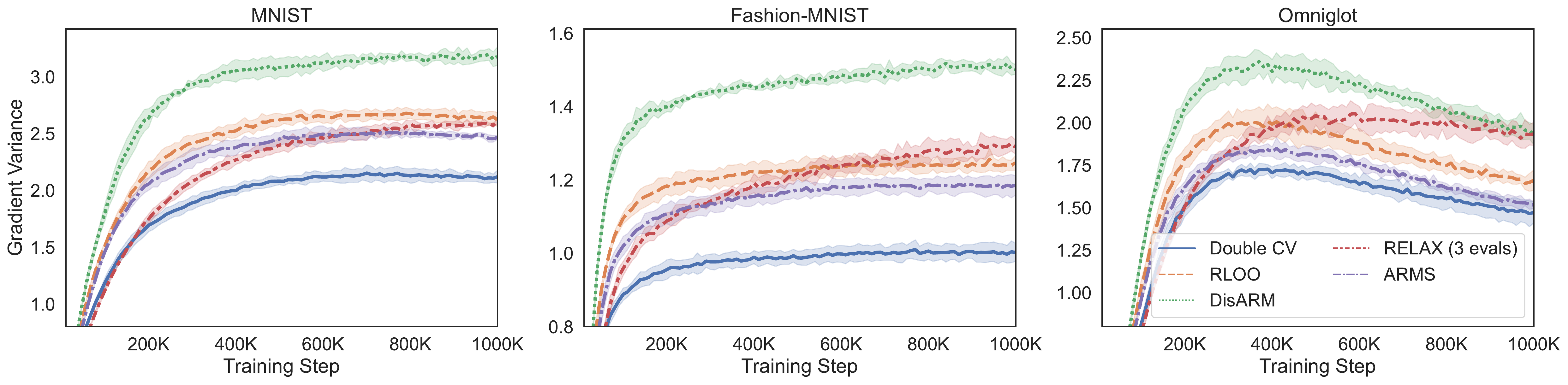}}
\end{tabular}
\vspace{-2mm}
\caption{
Variance of gradient estimates in training nonlinear binary latent variational autoencoders with $K=4$ (except RELAX which needs 3 evaluations of $f$) on MNIST, Fashion-MNIST, and Omniglot.
\emph{Top:} Using Bernoulli likelihoods and dynamically binarized datasets.
\emph{Bottom:} Using Gaussian likelihoods and non-binarized datasets.
} 
\label{fig:variance-nonlinear-K4}
\end{figure*}

We compared the following estimators: RLOO,  DisARM %
and the proposed Double CV method where all three estimators use $K$ samples. 
We experimented with $K=2$ and $K=4$.
For $K=4$ we also compare to the 
state-of-the-art ARMS estimator recently proposed by \citet{dimitrievzhou21}. 
Besides, we include in the comparison the RELAX estimator that combines concrete relaxation~\citep{Tucker2017} with 
a learned control variate 
\citep{Grathwohl2018}. 
We point out that RLOO, DisARM, Double CV, and ARMS (when $K=4$) have roughly the same running time on a P100 GPU while RELAX is computationally more expensive and 
is roughly twice slower than the other four estimators with $K=4$ (see Table~\ref{tab:time} in the Appendix). 
Also note that RELAX is less generally applicable since it assumes the existence of a continuous relaxation for $x$. 

\paragraph{Results}
Table~\ref{tab:K2} shows the training ELBO for binarized and continuous datasets when training the VAE 
by different estimators with $K=2$. 
We can observe that Double CV consistently outperforms 
RLOO in all experiments, while having approximately the same running 
time.
Double CV also outperforms 
DisARM in all cases for both Bernoulli and 
Gaussian likelihoods. 
Furthermore, Fig.~\ref{fig:dyn-nonlinear-K2} 
plots the gradient 
variance and the training ELBO for the binarized datasets
as a function of the training steps.  Similarly, 
Fig.~\ref{fig:cont-nonlinear-K2}
shows the corresponding results for the non-binarized (continuous) datasets where a Gaussian likelihood is used. 
We observe that the Double CV estimator can have lower
variance than RLOO and DisARM. 
Also, while RELAX performs better than the other methods it is less generally applicable and more expensive. 

For $K=4$, the final training ELBO values are reported in Table~\ref{tab:K4} and the variances 
of the different estimators are plotted in Fig.~\ref{fig:variance-nonlinear-K4}. 
We can observe 
that Double CV consistently has lower variance than other estimators
and it outperforms ARMS in terms of training ELBO in most cases. 
It also significantly outperforms RELAX. 
Note that, even with $K=4$, Double CV is still nearly twice faster than RELAX.

\section{CONCLUSION
\label{sec:conclusion}
}

We presented
a new variance 
reduction technique called double control variates for gradient estimation of discrete latent variable models.
We achieved substantial variance reduction by constructing control variates on top of existing leave-one-out baselines in REINFORCE estimators.
The proposed estimator is unbiased and adds no extra computational cost to the standard backpropagation cost needed for obtaining gradients over model parameters. 

Finally,  the use of double control variates can be orthogonal to other 
techniques for variance reduction such as coupled sampling \citep{yin2018arm,disarm,dongetal2021,dimitrievzhou21} and concrete relaxations \citep{Tucker2017,Grathwohl2018}. This could lead to various combinations of our approach  with these techniques. For instance, if we start from our initial estimator 
in \eqref{eq:doublecv_initial} where 
we simply replace the initial 
objective function $f(x)$ with the new effective objective $f(x) + \alpha b(x)$, a combination with coupled sampling is possible, e.g.\ certainly this holds for the mean field choice $b(x) = \nabla f(\mu)^\top (x - \mu)$. Also if we relax the restriction  
of the global correction $\Exp_{q_\eta(x)}[b(x)\nabla_\eta \log q_\eta (x)]$ to be analytic but instead allow to be reparametrizable, then our method could be combined with the concrete relaxation methods. The investigation of such combinations is an interesting topic for future research.

\bibliography{biblio}
\bibliographystyle{apalike}

\onecolumn

\appendix
\section{Proofs} 

\subsection{Proof of Proposition 1
} 

The RLOO estimator can be written as
\begin{align}
\underbrace{\frac{1}{K}\sum_{k=1}^K \left(f(x_k) - \Exp f\right)\nabla_\eta\log q_\eta(x_k)}_{\text{R}^*}
+ \underbrace{\frac{1}{K}\sum_{k=1}^K \left(\Exp f - \frac{1}{K-1}\sum_{
   j\neq k} f(x_j)\right)\nabla_\eta\log q_\eta(x_k)}_{E}  
\label{eq:loogeneral2}   
\end{align}
where $\text{R}^*$ is the REINFORCE estimator with baseline $\Exp f$ and $E$ is a residual term of zero mean.  
To prove the Proposition we will use 
$\Var(\text{RLOO}) = \Var(\text{R}^* + E) = \Var(\text{R}^*) + \Var(E) + 2 \Cov(\text{R}^*,E)$. Then, it suffices to show that $\Cov(\text{R}^*,E)=0$.
We have 
$$
\Cov(\text{R}^*,E) 
= \frac{1}{K^2} 
\sum_{k=1}^K \sum_{k'=1}^K 
\Exp \left[(f(x_k) - \Exp f) 
(\Exp f -  f_{-k'}) \nabla_\eta\log q_\eta(x_{k}) \nabla_\eta\log q_\eta(x_{k'})^\top \right]
$$
where we used $f_{-k'} = \frac{1}{K-1}\sum_{
   j\neq k'} f(x_j)$ for short. For all terms in the double sum such that $k=k'$ the expectation
$$   
\Exp \left[(f(x_k) - \Exp f) 
(\Exp f -  f_{-k}) \nabla_\eta\log q_\eta(x_{k}) \nabla_\eta\log q_\eta(x_{k})^\top \right]= 0   
$$   
because the zero-mean random variable $\Exp f -  f_{-k}$ is independent from the remaining product (since it does not contain the sample $x_k$). For all cross terms $k \neq k'$
the whole product $(f(x_k) - \Exp f) 
(\Exp f -  f_{-k'}) \nabla_\eta\log q_\eta(x_{k})$ does not contain the sample $x_{k'}$. Therefore this product
is independent from $\nabla_\eta\log q_\eta(x_{k'})$
and thus each cross term is zero because of the 
score function property 
$\Exp[ \nabla_\eta\log q_\eta(x_{k'})]=0$.
This shows that $\Cov(\text{R}^*,E)=0$
which completes the proof. 

\subsection{Proof of Proposition 2}

It holds
\begin{align}
& \Exp_{q_\eta(x_{1:K})} 
[\gamma_k(x_{1:K}) \nabla_\eta \log q_\eta(x_k)] \nonumber \\
& = \Exp_{q_\eta(x_k)} 
\bigg[ 
\Exp_{q_\eta(x_{1:k-1},x_{k+1:K})}
[\gamma_k(x_{1:K}) ]
\nabla_\eta \log q_\eta(x_k) \bigg] 
\nonumber \\
& = \Exp_{q_\eta(x_k)} 
\bigg[ \text{const}
\nabla_\eta \log q_\eta(x_k) \bigg] 
= 0.
\end{align}
where the last line is just a consequence of the score function 
property since  $\text{const}$ does not depend on $x_k$. 

\subsection{Proof of Proposition 3}

The estimator can be written as

\begin{align}
& \frac{1}{K} \sum_{k=1}^K \left[f(x_k)  -  \frac{1}{K - 1}\sum_{j\neq k} f(x_j) \right] \nabla_{\eta}  \log q_\eta(x_{k}) \nonumber \\
& + \alpha \frac{1}{K} \sum_{k=1}^K \left( 
\textcolor{blue}{b_k(x_{1:K})} - \frac{1}{K - 1}\sum_{j\neq k} \textcolor{red}{b_j(x_{1:K})} \right) \nabla_{\eta} \log q_\eta(x_{k}) \nonumber \\
& - \alpha \Exp_{q(x)}[\nabla_\eta \log q_\eta(x) \times (x-\mu)^\top]
    \left( \frac{1}{K}\sum_{k=1}^K \nabla f(x_k) \right),
\label{eq:doublyEstimator_app} 
\end{align}
where 
$b_k(x_{1:K}) = \left( \frac{1}{K-1}
\sum_{ j \neq k}
\nabla f(x_j) \right)^\top (x_k - \mu)
$ and $b_j(x_{1:K}) = \left( \frac{1}{K-1}
\sum_{ m \neq j}
\nabla f(x_m) \right)^\top (x_j - \mu)$. 
It suffices to show that the expectation of the second line is minus the correction term at the 
third line. The expectation of each term $b_j(x_{1:K}) \nabla_{\eta} \log q_\eta(x_{k})$ for $j \neq k$ is zero because the zero-mean term $x_j - \mu$ is always 
independent from the rest of the terms in the product. 
Then, we need to examine only the expectation of 
$$
\frac{1}{K} \sum_{k=1}^K 
\textcolor{blue}{b_k(x_{1:K})} \nabla_{\eta} \log q_\eta(x_{k}) =
\frac{1}{K (K-1)} \sum_{k=1}^K 
  \nabla_{\eta} \log q_\eta(x_{k})  (x_k - \mu)^\top  
 \sum_{j \neq k} \nabla f(x_j).
$$
Then observe  that the expectation of $\nabla_{\eta} \log q_\eta(x_{k}) \times (x_k - \mu)^\top$ is the same 
for every sample $x_k$, so the above reduces to 
$$
\Exp_{q_\eta(x)}[\nabla_\eta \log q_\eta(x) \times (x-\mu)^\top]
\frac{1}{K (K-1)} \sum_{k=1}^K 
 \sum_{j \neq k} \nabla f(x_j)
$$
from which the result follows since 
$\sum_{k=1}^K 
 \sum_{j \neq k} \nabla f(x_j) = (K-1) \sum_{k=1}^K \nabla f(x_k)$.

\subsection{The Optimal Value of $\alpha$ for $K=2$} 

The gradient for $K=2$ 
can be written as 
{\small
\begin{align}
& \frac{1}{2}[f(x_1) - f(x_2)]  
(\nabla_\eta \log q_\eta(x_1) - \nabla_\eta \log q_\eta(x_2)) \nonumber \\
& - 
\frac{1}{2}
\alpha\left( M (
\nabla f(x_1) + \nabla f(x_2))
- [\nabla f(x_2)^\top (x_1 -\mu) - \nabla f(x_1)^\top (x_2 - \mu)](
\nabla_\eta \log q_\eta(x_1) - \nabla_\eta \log q_\eta(x_2)) 
\right)
\label{eq:doublyEstimatorK=2app} 
\end{align}
}

where $M = \Exp_{q_\eta(x)}[\nabla_\eta \log q_\eta(x) \times (x-\mu)^\top]$.
If we denote 
$$
g(x_1,x_2) = 
[f(x_1) - f(x_2)]  
(\nabla_{\eta} \log q_\eta(x_1) - \nabla_\eta \log q_\eta(x_2))
$$ 
and 
{ \small
$$
h(x_1,x_2)  
= M (
\nabla f(x_1) + \nabla f(x_2))
- [\nabla f(x_2)^\top (x_1 -\mu) - \nabla f(x_1)^\top (x_2 - \mu)](
\nabla_\eta \log q_\eta(x_1) - \nabla_\eta \log q_\eta(x_2)) 
$$
}

the gradient can be written as 
$$
\frac{1}{2} 
\left( g(x_1, x_2)  - \alpha h(x_1, x_2) \right). 
$$
Then the optimal $\alpha$ that minimizes the variance is given by 
$$
\alpha = \frac{\Exp [g(x_1,x_2)^\top h(x_1, x_2)]}
{\Exp [h(x_1,x_2)^\top h(x_1,x_2)]}
$$
Similarly we can construct the optimal value of 
$\alpha$ for any $K > 2$.

\subsection{The ``half'' Double Control
Variate Estimators} 

\begin{figure}[ht]
\centering
\begin{tabular}{cc}
{\includegraphics[scale=0.35]
{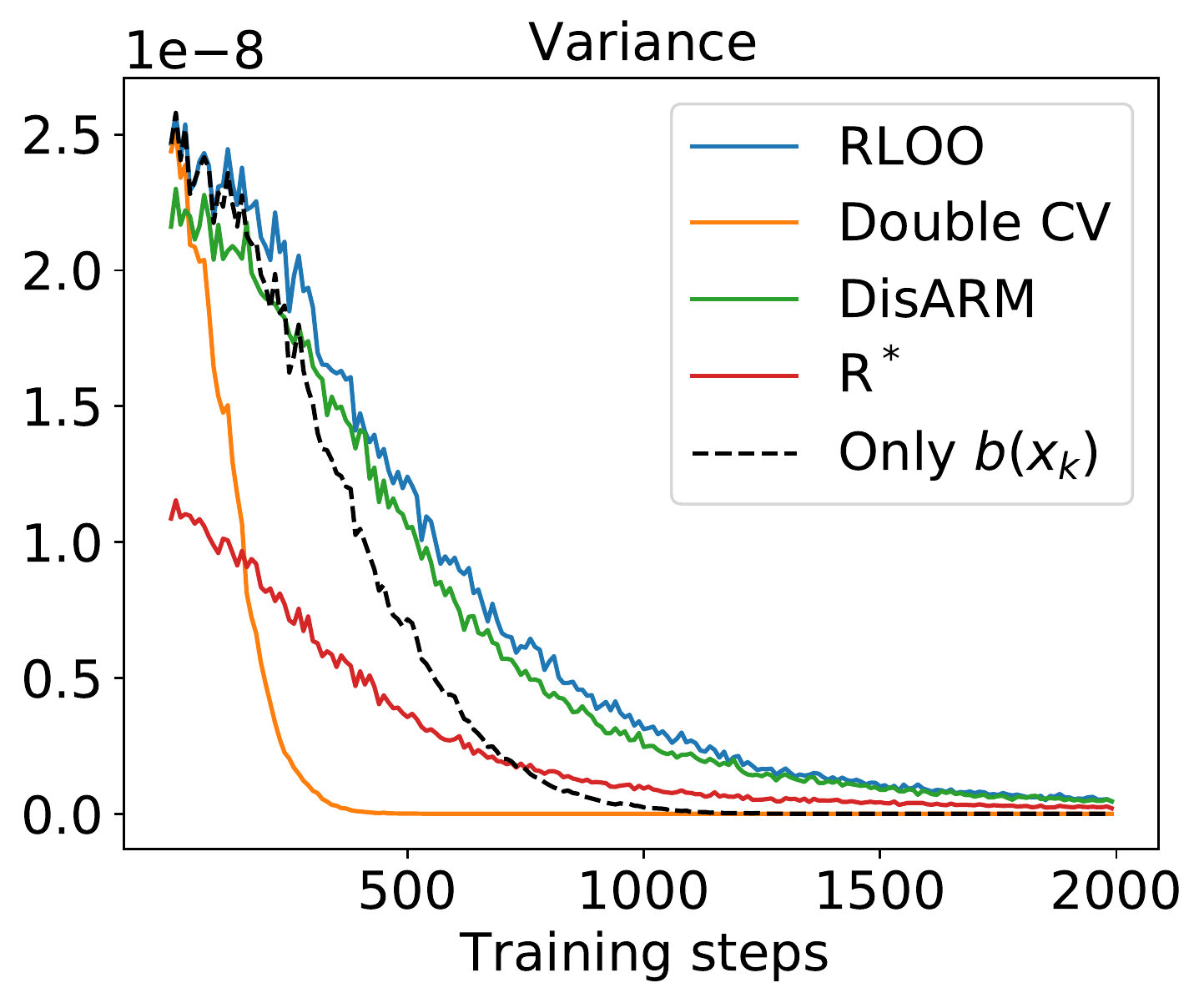}} &
{\includegraphics[scale=0.35]
{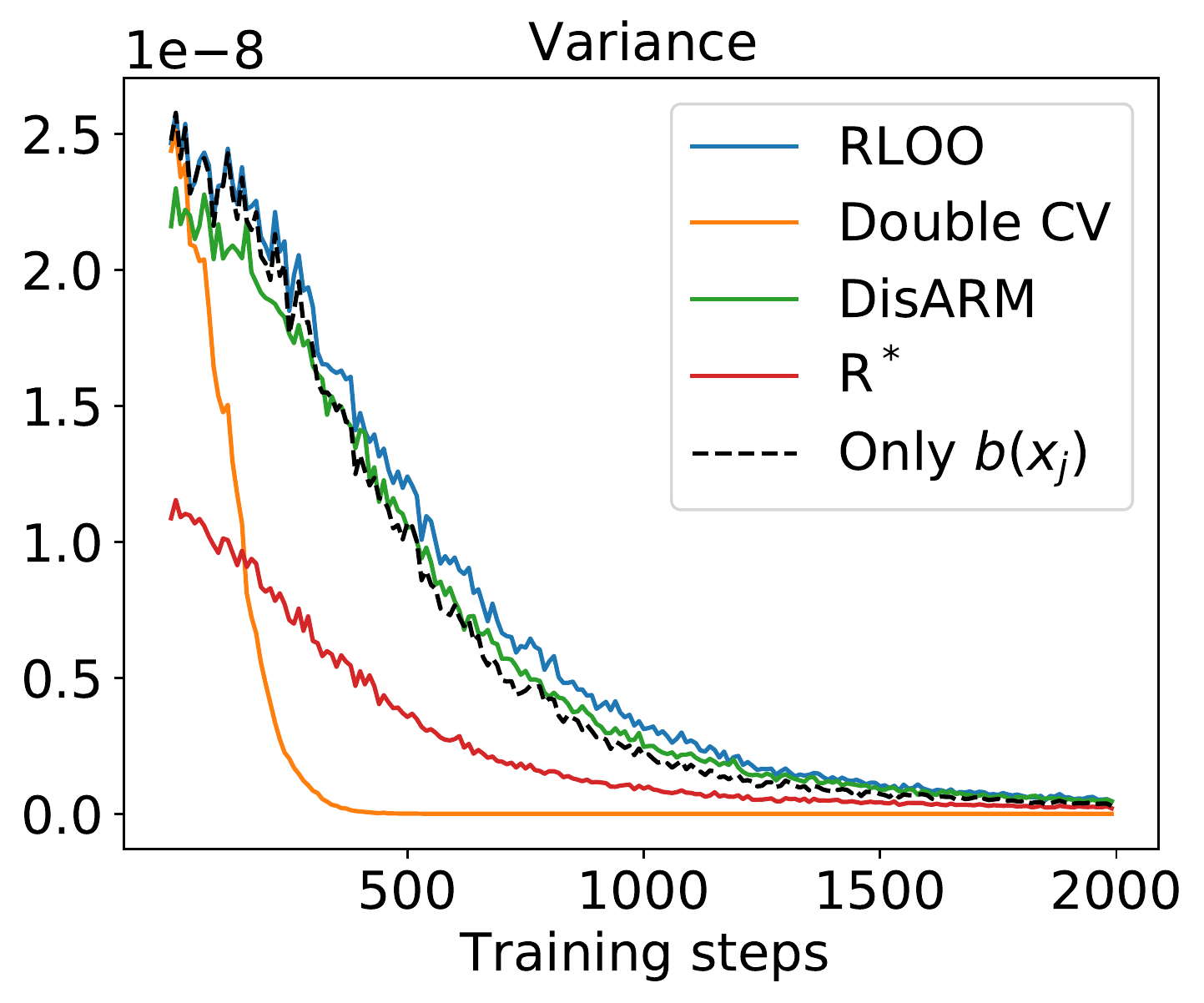}}
\end{tabular}
\caption{
\emph{Left:} Variance of the  ``only $b(x_k)$'' estimator where only the half part of the double control variate is used. \emph{Right:} The corresponding plot 
for the ``only $b(x_j)$'' estimator
where the other half part of the double control variate is used. The full 
double control variate estimator  
(Double CV), RLOO, DisARM and $\text{R}^*$ are included for comparison. The experiment corresponds to the toy problem with $D=10$ and $b(x)$ was chosen according to Eq.\ \eqref{eq:bk},
i.e.\ the full Double CV estimator is from  \eqref{eq:doublyEstimator3}.
} 
\label{fig:toy_exampleD1_onlyonebx_app}
\end{figure}

One question is whether we need both $b(x_k)$ and 
$b(x_j)$ or we could keep one of them, i.e.\ to use an ``$b(x_k)$ only'' or ``$b(x_j)$ only'' estimator. It is straightforward to express these latter unbiased estimators, 
as follows. The ``$b(x_k)$ only'' estimator is given by 
\begin{equation}
    \frac{1}{K} \! \! \sum_{k=1}^K \! \! \left[f(x_k) + \alpha \textcolor{blue}{b(x_k)} \! - \!  \frac{1}{K \! - \! 1}\sum_{j\neq k} \! f(x_j) \right] \!
     \nabla_{\eta}  \log q_\eta(x_{k}) -  \alpha \Exp_{q_\eta(x)} [b(x) \nabla_{\eta} \! \log q_\eta(x)].
\label{eq:doublyEstimator_onlybxk} 
\end{equation}
and the ``$b(x_j)$ only'' by 
\begin{equation}
    \frac{1}{K} \! \! \sum_{k=1}^K \! \! \left[f(x_k) \! - \!  \frac{1}{K \! - \! 1}\sum_{j\neq k} \! (f(x_j)  + \alpha \textcolor{red}{b(x_j)} ) \right] \!
     \nabla_{\eta}  \log q_\eta(x_{k}).
\label{eq:doublyEstimator_onlybxj} 
\end{equation}
It is easy to show that both estimators are unbiased. 
However, in practice these estimators can be much less effective in terms of variance reduction than their Double CV combination. In Fig.\
\ref{fig:toy_exampleD1_onlyonebx_app}
we apply these two estimators to the toy learning problem with $D=10$. 
Both estimators are significantly outperformed by the full Double CV estimator. Notably, the ``$b(x_k)$ only'' estimator could outperform 
$\text{R}^*$ since it uses a baseline
that depends on the current sample
$x_k$, while ``$b(x_j)$ only''
reduces the variance of the RLOO 
control variate but remains bounded by 
$\text{R}^*$.

\section{Additional Results}

\subsection{Toy Experiment with $D=1$}

\begin{figure}[ht]
\centering
\begin{tabular}{cc}
{\includegraphics[scale=0.35]
{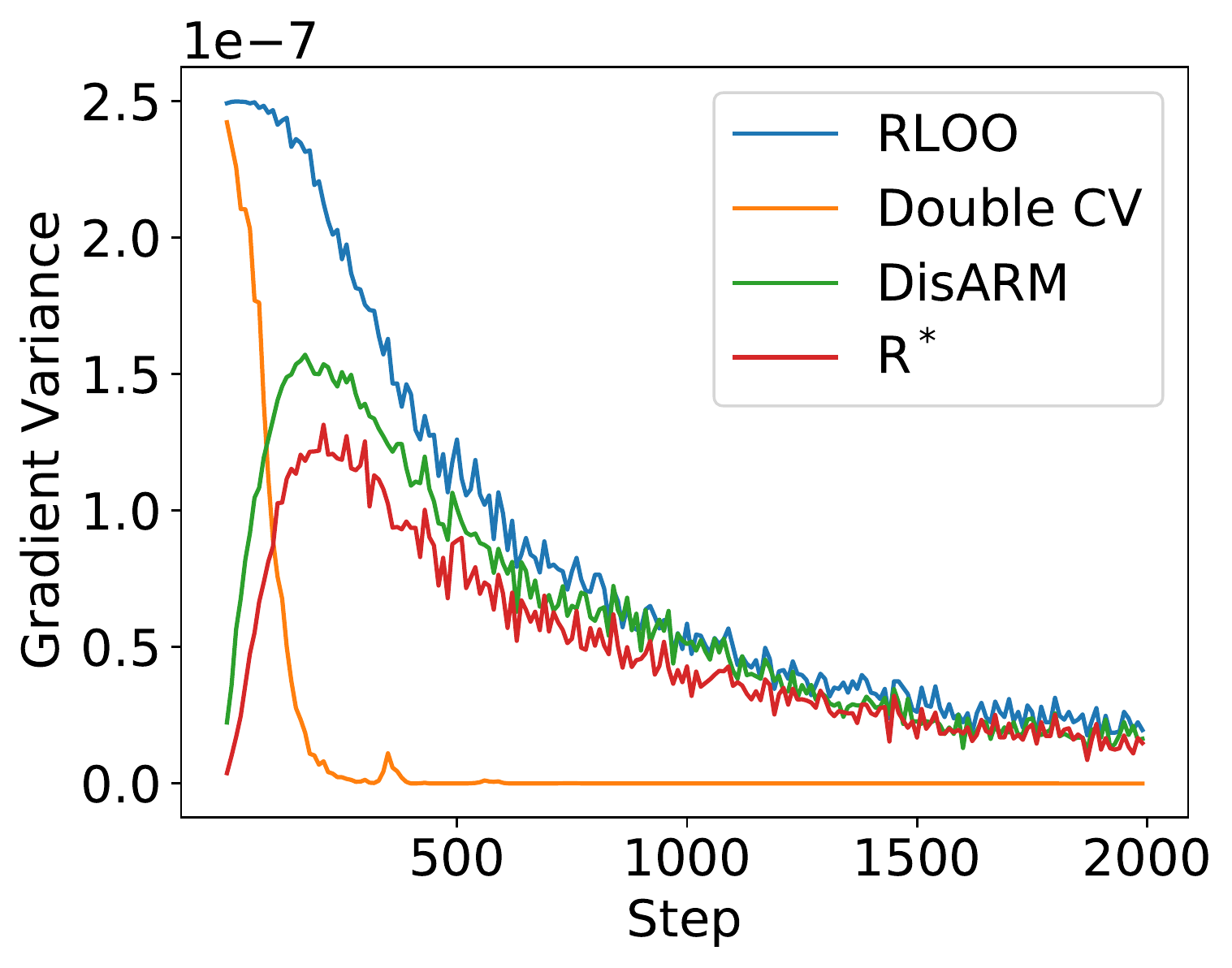}} &
{\includegraphics[scale=0.35]
{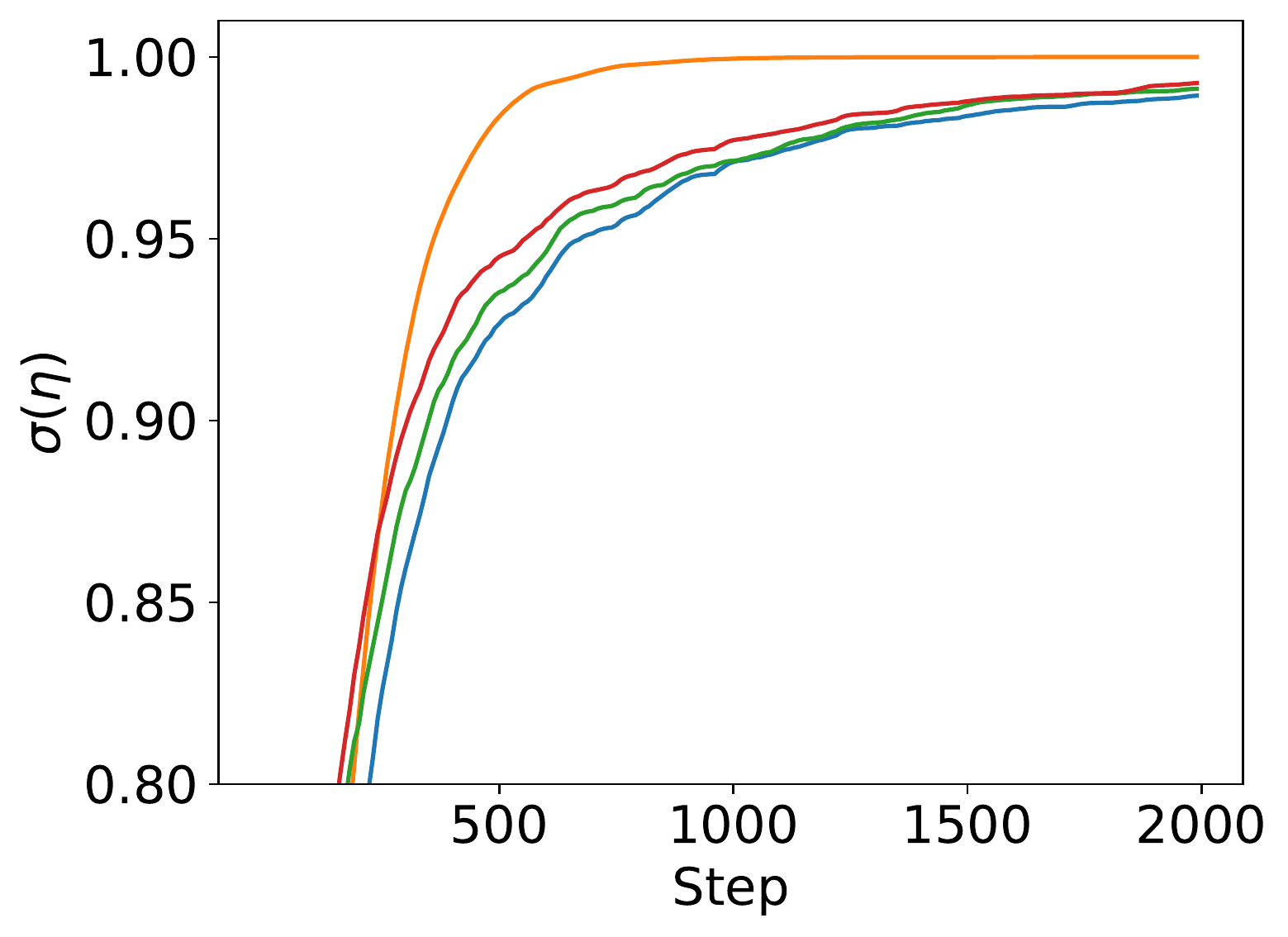}}
\end{tabular}
\caption{\emph{Left:} Variance of the  gradient estimators for the toy problem with $D=1$. \emph{Right:} The estimated value $\sigma(\eta)$ across iterations (optimal value is $1$).  
} 
\label{fig:toy_exampleD1}
\end{figure}

\begin{figure}[ht]
\centering
\begin{tabular}{ccc}
{\includegraphics[scale=0.3]
{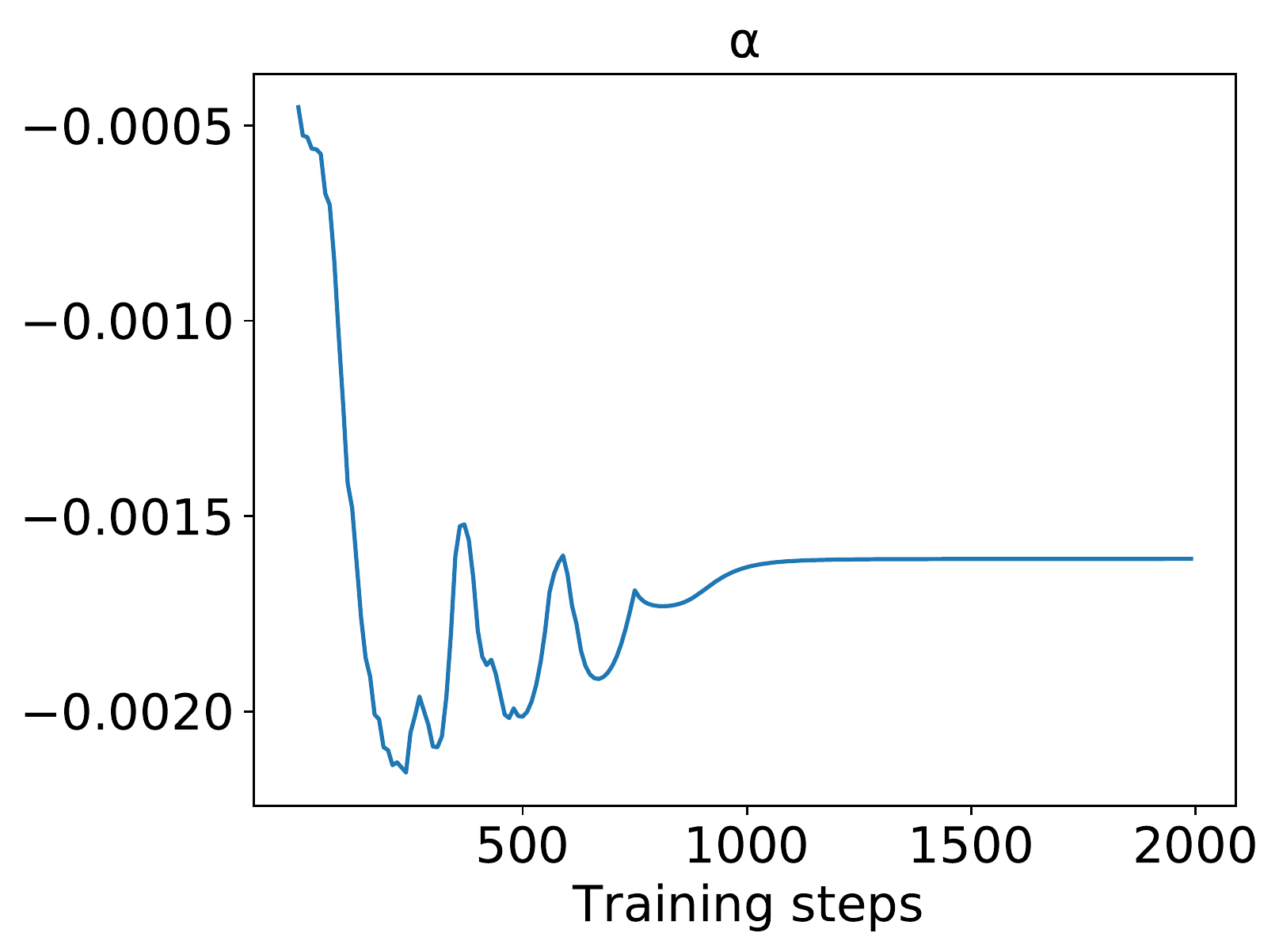}} &
{\includegraphics[scale=0.3]
{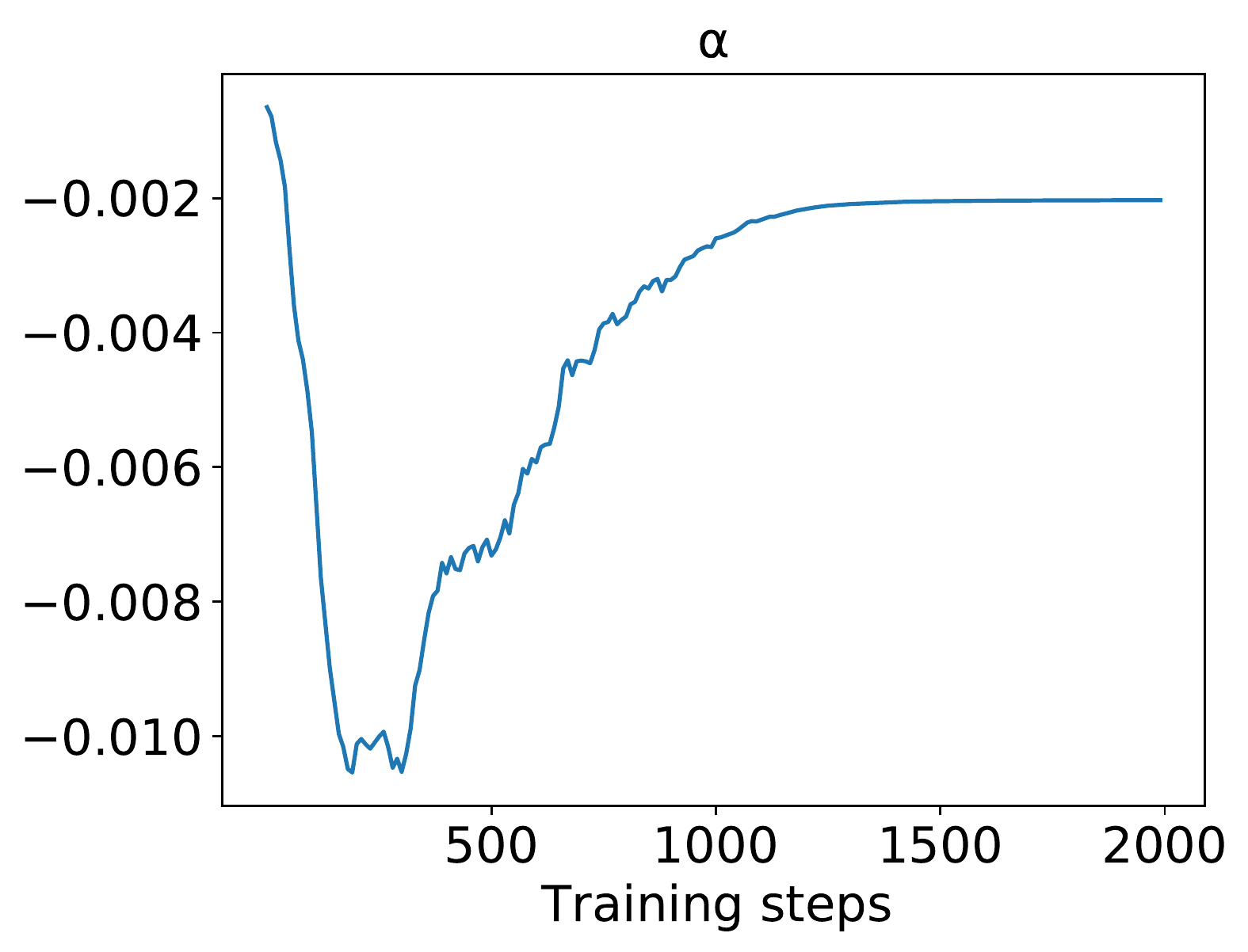}}
& 
{\includegraphics[scale=0.3]
{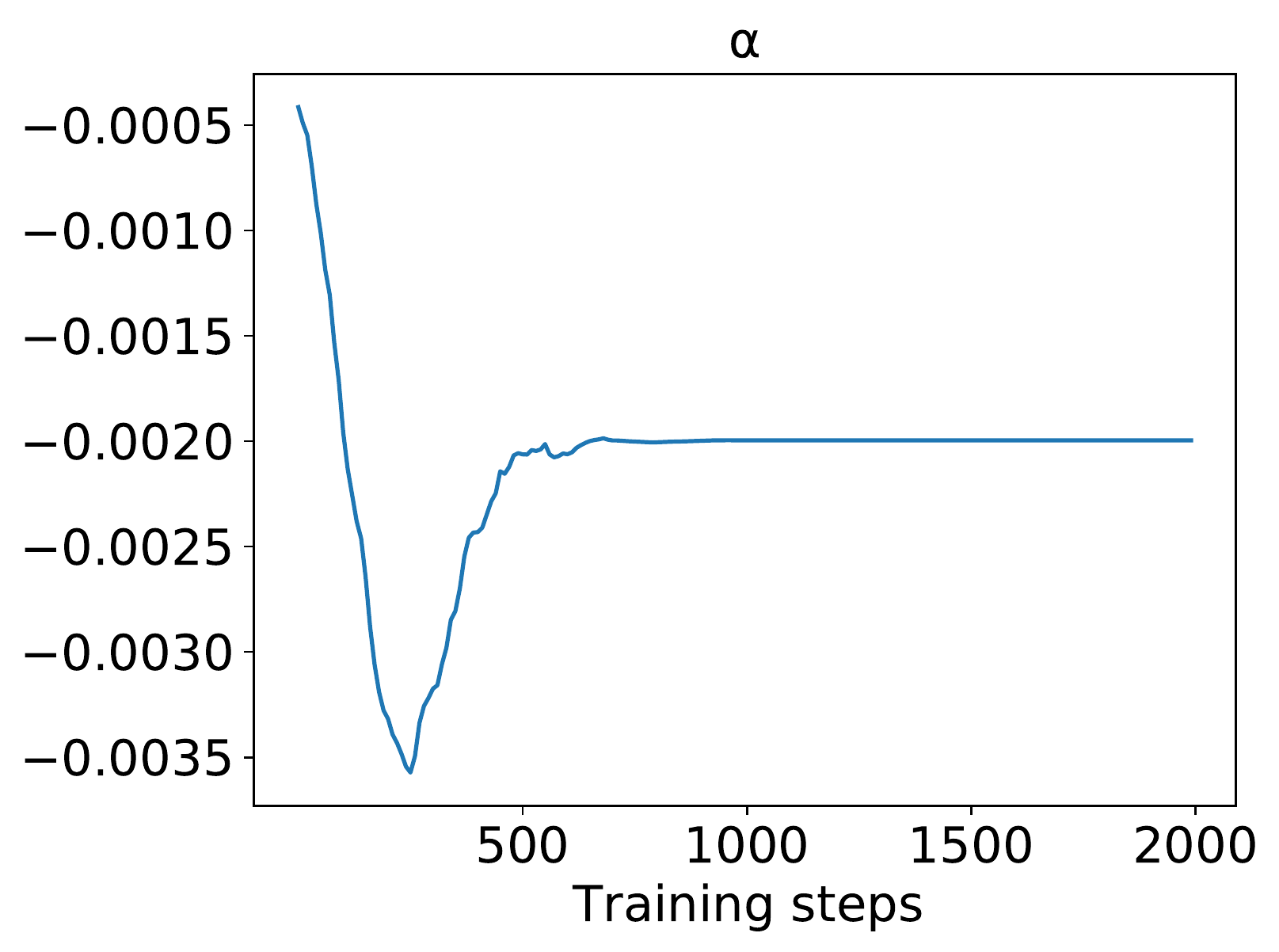}}
\\
$D=1$ & $D=10$  & $D=200$ 
\end{tabular}
\caption{
The evolution of the estimated regression coefficient $\alpha$ during optimization for the toy learning problem. 
} 
\label{fig:toy_alpha}
\end{figure}

For completeness, we include the results of a simpler version of the toy experiment described in Section~\ref{sec:toyproblem}, where we set $D=1$.
This is the setting used in several previous works~\citep{Tucker2017,Grathwohl2018,yin2018arm,disarm}. 
The variances of the gradient estimators and the training curves of $\sigma(\eta)$ are plotted in Fig.~\ref{fig:toy_exampleD1}.
Fig.~\ref{fig:toy_alpha} shows the 
evolution of the estimated regression coefficient $\alpha$.

\subsection{Training Binary Latent VAEs}

\subsubsection{Time comparison}

In Fig.~\ref{tab:time} we report the per-step running time of RLOO, Double CV, DisARM, ARMS estimators when $K=4$ and compare to RELAX. 
RELAX is almost twice slower.

\begin{table}[ht]
\centering
\begin{tabular}{lrrrrr}
\toprule
& \multicolumn{1}{c}{RLOO} & \multicolumn{1}{c}{Double CV}  & \multicolumn{1}{c}{DisARM} & \multicolumn{1}{c}{ARMS} & \multicolumn{1}{c}{RELAX} \\
\midrule
Time (sec/step) & 0.0035 & 0.0036 & 0.0031 & 0.0037 & 0.0080 \\
\bottomrule
\end{tabular}
\caption{Time per step when training a Bernoulli VAE with $K=4$ (except RELAX which needs 3 evaluations of $f$) on dynamically binarized Fashion-MNIST.}
\label{tab:time}
\end{table}

\begin{table*}[t]
\footnotesize
\centering
\begin{tabular}{lrrr|r}
\toprule
& \multicolumn{1}{c}{RLOO} & \multicolumn{1}{c}{Double CV}  & \multicolumn{1}{c}{DisARM} & \multicolumn{1}{c}{RELAX (3 evals)} \\
\midrule
\multicolumn{5}{l}{\emph{MNIST}:} \\
Linear
&
$-113.06\pm0.05$
&
$-112.82\pm0.07$
&
$\bf -112.72\pm0.07$
&
$\bf -112.18\pm0.07$
\\
Nonlinear
&
$-103.11\pm0.16$
&
$\bf -102.45\pm0.13$
&
$-102.56\pm0.09$
&
$\bf -101.86\pm0.11$
\\
\midrule
\multicolumn{5}{l}{\emph{Fashion-MNIST}:} \\
Linear
&
$-257.38\pm0.17$
&
$\bf -256.21\pm0.17$
&
$-257.01\pm0.06$
&
$\bf -255.16\pm0.17$
\\
Nonlinear
&
$-241.53\pm0.24$
&
$\bf -240.96\pm0.17$
&
$-241.02\pm0.20$
&
$\bf -240.63\pm0.16$
\\
\midrule
\multicolumn{5}{l}{\emph{Omniglot}:} \\
Linear
&
$-119.63\pm0.05$
&
$-119.52\pm0.02$
&
$\bf -119.42\pm0.03$
&
$\bf -119.16\pm0.02$
\\
Nonlinear
&
$-116.83\pm0.05$
&
$\bf -116.22\pm0.08$
&
$-116.36\pm0.05$
&
$-\bf 115.79\pm0.06$
\\
\bottomrule
\end{tabular}
\caption{Training binary latent VAEs with $K=2$ (except RELAX which needs 3 evaluations of $f$) on dynamically binarized MNIST, Fashion-MNIST, and Omniglot. We report the average ELBO on the training set over 5 independent runs.}
\label{tab:K2-bin}
\end{table*}

\begin{table*}[t]
\footnotesize
\centering
\begin{tabular}{lrrr|r}
\toprule
& \multicolumn{1}{c}{RLOO} & \multicolumn{1}{c}{Double CV}  & \multicolumn{1}{c}{DisARM} & \multicolumn{1}{c}{RELAX (3 evals)} \\
\midrule
\multicolumn{5}{l}{\emph{MNIST}} \\
Linear
&
$503.01\pm0.22$
&
$504.33\pm0.98$
&
$\bf 504.43\pm0.93$
&
$\bf 513.38\pm0.52$
\\
Nonlinear
&
$668.07\pm0.40$
&
$\bf 676.87\pm1.18$
&
$668.03\pm0.61$
&
$\bf 688.58\pm0.52$
\\
\midrule
\multicolumn{5}{l}{\emph{Fashion-MNIST}} \\
Linear
&
$29.75\pm0.40$
&
$31.08\pm0.24$
&
$\bf 31.71\pm0.20$
&
$\bf 37.54\pm0.30$
\\
Nonlinear
&
$179.52\pm0.23$
&
$\bf 186.35\pm0.64$
&
$182.65\pm0.47$
&
$\bf 196.38\pm0.66$
\\
\midrule
\multicolumn{5}{l}{\emph{Omniglot}} \\
Linear
&
$245.73\pm0.33$
&
$245.97\pm1.02$
&
$\bf 247.70\pm0.85$
&
$\bf 255.69\pm0.70$
\\
Nonlinear
&
$443.51\pm0.93$
&
$\bf 446.95\pm0.63$
&
$446.22\pm1.38$
&
$\bf 462.30\pm0.91$
\\
\bottomrule
\end{tabular}
\caption{Training binary latent VAEs with Gaussian likelihoods using $K=2$ (except RELAX which needs 3 evaluations of $f$) on non-binarized MNIST, Fashion-MNIST, and Omniglot. We report the average ELBO on the training set over 5 independent runs.}%
\label{tab:K2-cont}
\end{table*}

\subsubsection{Full results of training ELBOs}

Here we include the full results of final training ELBOs from the experiment in Section~\ref{sec:vae}. 
Table~\ref{tab:K2-bin} and Table~\ref{tab:K2-cont} extend Table~\ref{tab:K2} to include the linear VAE results trained under the same setting. 
Table~\ref{tab:K4-bin} and Table~\ref{tab:K4-cont} extend Table~\ref{tab:K4} to include the linear VAE results trained under the same setting. 
The linear VAE has $200$ dimensional latent variable $x$ and use a single fully-connected layer to produce the logits (for Bernoulli likelihoods) or the mean (for Gaussian likelihoods) of the distribution of $y$.

\begin{table*}[t]
\footnotesize
\centering
\begin{tabular}{lrrrr}
\toprule
& \multicolumn{1}{c}{RLOO} & \multicolumn{1}{c}{Double CV}  & \multicolumn{1}{c}{DisARM} & \multicolumn{1}{c}{ARMS} \\
\midrule
\multicolumn{5}{l}{\emph{MNIST}:} \\
Linear
&
$-111.89\pm0.09$
&
$\bf -111.79\pm0.09$
&
$-112.01\pm0.06$
&
$-111.87\pm0.02$
\\
Nonlinear
&
$-100.50\pm0.22$
&
$\bf -99.89\pm0.12$
&
$-100.67\pm0.07$
&
$-100.07\pm0.08$
\\
\midrule
\multicolumn{5}{l}{\emph{Fashion-MNIST}:} \\
Linear
&
$-254.59\pm0.16$
&
$\bf -254.52\pm0.23$
&
$-255.01\pm0.10$
&
$-254.67\pm0.20$
\\
Nonlinear
&
$-239.03\pm0.15$
&
$-238.98\pm0.18$
&
$-239.20\pm0.15$
&
$\bf -238.50\pm0.13$
\\
\midrule
\multicolumn{5}{l}{\emph{Omniglot}:} \\
Linear
&
$-118.89\pm0.02$
&
$-118.95\pm0.02$
&
$-118.97\pm0.01$
&
$\bf -118.87\pm0.02$
\\
Nonlinear
&
$-114.75\pm0.07$
&
$\bf -114.56\pm0.06$
&
$-115.05\pm0.07$
&
$-114.57\pm0.06$
\\
\bottomrule
\end{tabular}
\caption{Training binary latent VAEs with $K=4$ on dynamically binarized MNIST, Fashion MNIST, and Omniglot. We report the average ELBO on the training set over 5 independent runs.}
\label{tab:K4-bin}
\end{table*}

\begin{table*}[ht]
\footnotesize
\centering
\begin{tabular}{lrrrr}
\toprule
& \multicolumn{1}{c}{RLOO} & \multicolumn{1}{c}{Double CV}  & \multicolumn{1}{c}{DisARM} & \multicolumn{1}{c}{ARMS} \\
\midrule
\multicolumn{5}{l}{\emph{MNIST}:} \\
Linear
&
$\bf 516.65\pm0.54$
&
$515.79\pm0.71$
&
$512.47\pm0.72$
&
$514.55\pm0.71$
\\
Nonlinear
&
$687.83\pm0.50$
&
$\bf 691.51\pm0.75$
&
$683.28\pm0.89$
&
$687.26\pm1.21$
\\
\midrule
\multicolumn{5}{l}{\emph{Fashion-MNIST}:} \\
Linear
&
$36.70\pm0.41$
&
$36.61\pm0.34$
&
$34.90\pm0.52$
&
$\bf 37.56\pm0.43$
\\
Nonlinear
&
$195.27\pm0.24$
&
$\bf 199.01\pm0.60$
&
$192.96\pm0.29$
&
$197.25\pm0.48$
\\
\midrule
\multicolumn{5}{l}{\emph{Omniglot}:} \\
Linear
&
$257.43\pm0.16$
&
$257.88\pm0.69$
&
$254.99\pm0.69$
&
$\bf 258.22\pm0.18$
\\
Nonlinear
&
$460.23\pm1.42$
&
$463.03\pm0.94$
&
$458.38\pm0.88$
&
$\bf 463.30\pm0.86$
\\
\bottomrule
\end{tabular}
\caption{Training binary latent VAEs with Gaussian likelihoods using $K=4$ on non-binarized MNIST, Fashion-MNIST, and Omniglot. We report the average ELBO on the training set over 5 independent runs.}
\label{tab:K4-cont}
\end{table*}

\subsubsection{Additional figures for nonlinear VAEs}

\begin{figure*}[ht]
\centering
\begin{tabular}{c}
{\includegraphics[width=0.96\textwidth]
{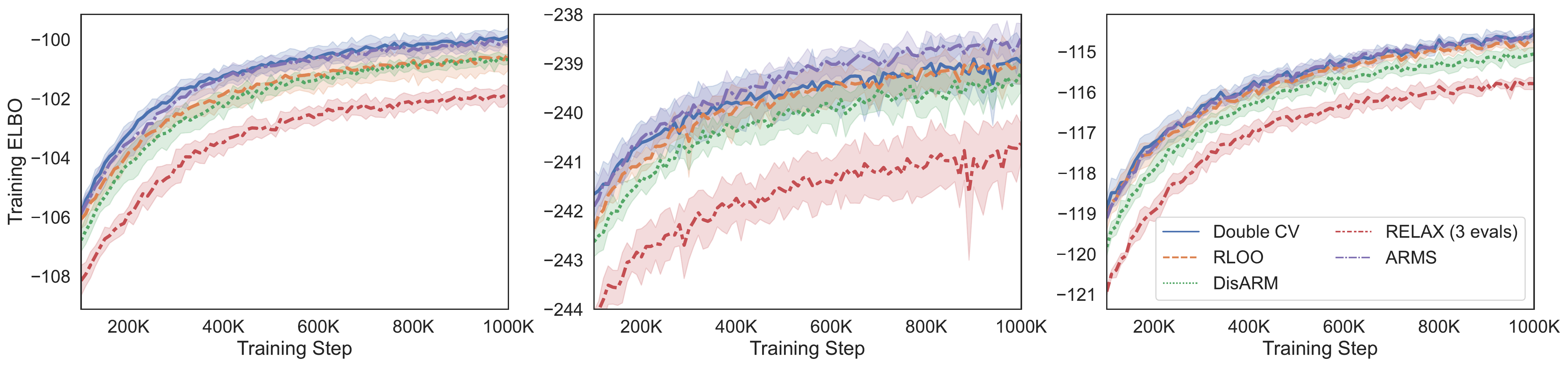}} \\
{\includegraphics[width=0.96\textwidth]
{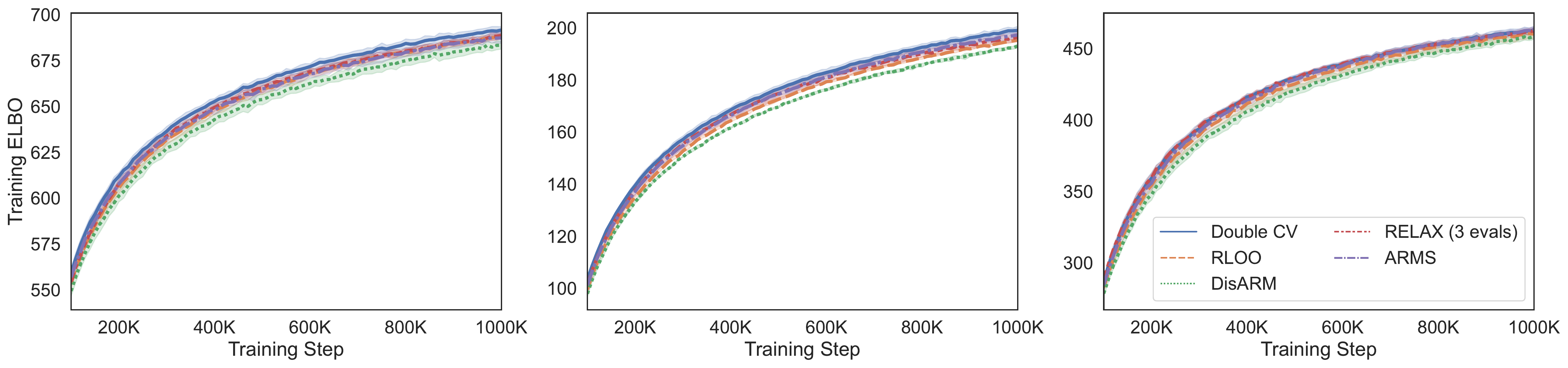}}
\end{tabular}
\caption{
Average training ELBOs for nonlinear binary latent VAEs trained by different estimators with $K=4$ (except RELAX which needs 3 evaluations of $f$) on MNIST, Fashion-MNIST, and Omniglot.
\emph{Top:} Using Bernoulli likelihoods and dynamically binarized datasets.
\emph{Bottom:} Using Gaussian likelihoods and non-binarized datasets.
} 
\label{fig:elbo-nonlinear-K4}
\end{figure*}

In Fig.~\ref{fig:elbo-nonlinear-K4} we plot the average training ELBOs as a function of training steps from the $K=4$ experiment in Section~\ref{sec:vae}.

\subsubsection{Additional figures for linear VAEs}

\begin{figure*}[ht]
\centering
\begin{tabular}{c}
{\includegraphics[width=0.96\textwidth]
{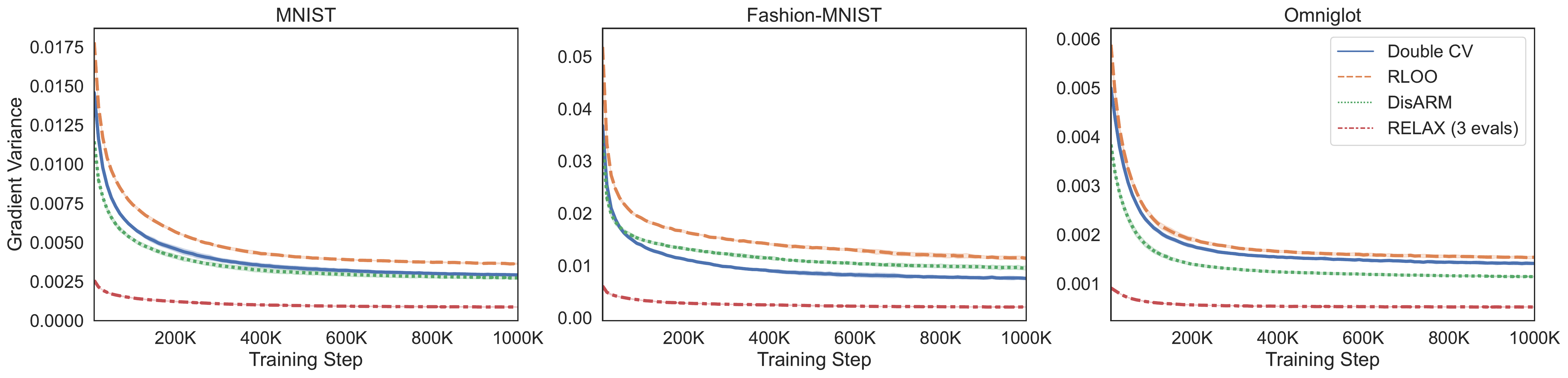}} \\
{\includegraphics[width=0.96\textwidth]
{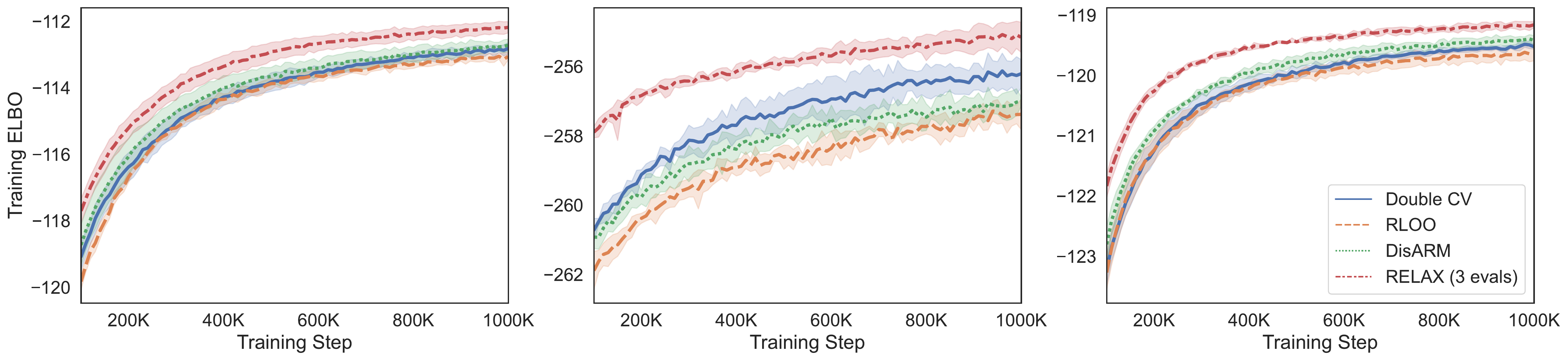}}
\end{tabular}
\caption{
Training linear binary latent VAEs with Bernoulli likelihoods with $K=2$ (except RELAX which needs 3 evaluations of $f$) on dynamically binarized MNIST, Fashion-MNIST, and Omniglot.
\emph{Top:} Variance of gradient estimates.
 \emph{Bottom:} Average ELBO on training examples.
} 
\label{fig:dyn-linear-K2}
\end{figure*}

\begin{figure*}[ht]
\centering
\begin{tabular}{c}
{\includegraphics[width=0.96\textwidth]
{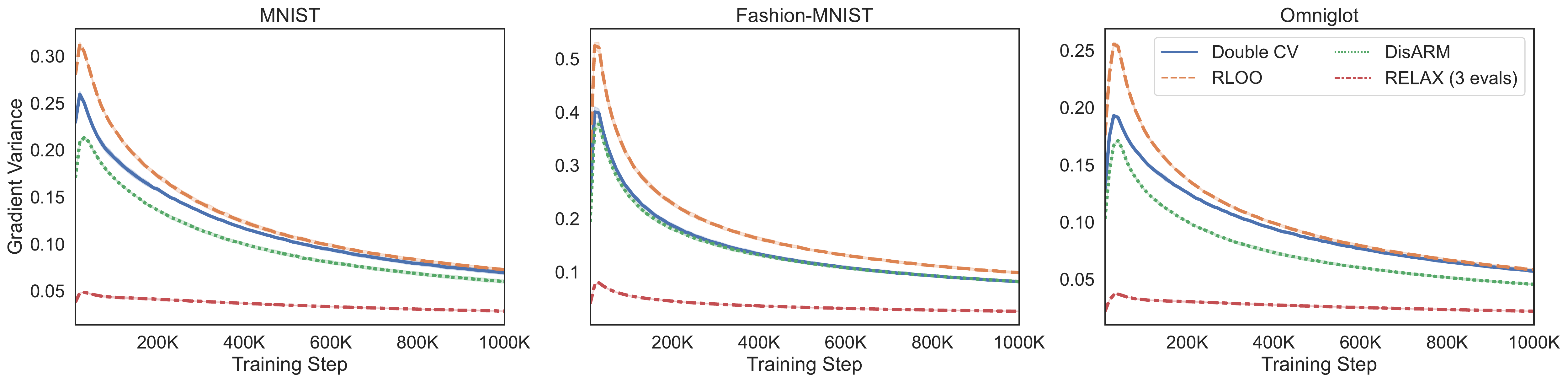}} \\
{\includegraphics[width=0.96\textwidth]
{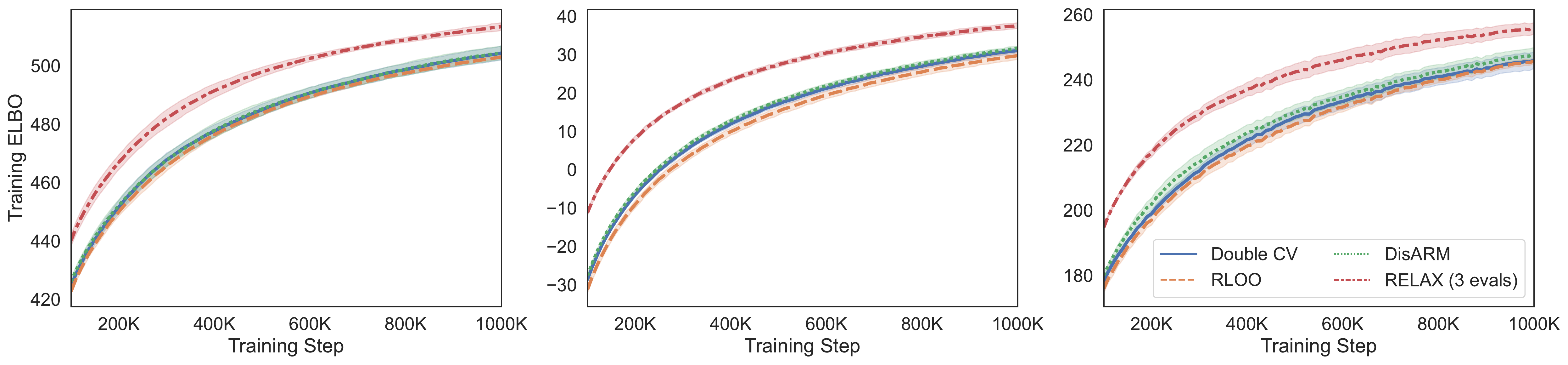}}
\end{tabular}
\caption{
Training linear binary latent VAEs with Gaussian likelihoods with $K=2$ (except RELAX which needs 3 evaluations of $f$) on non-binarized MNIST, Fashion-MNIST, and Omniglot.
\emph{Top:} Variance of gradient estimates.
 \emph{Bottom:} Average ELBO on training examples.
} 
\label{fig:cont-linear-K2}
\end{figure*}

\begin{figure*}[ht]
\centering
\begin{tabular}{c}
{\includegraphics[width=0.96\textwidth]
{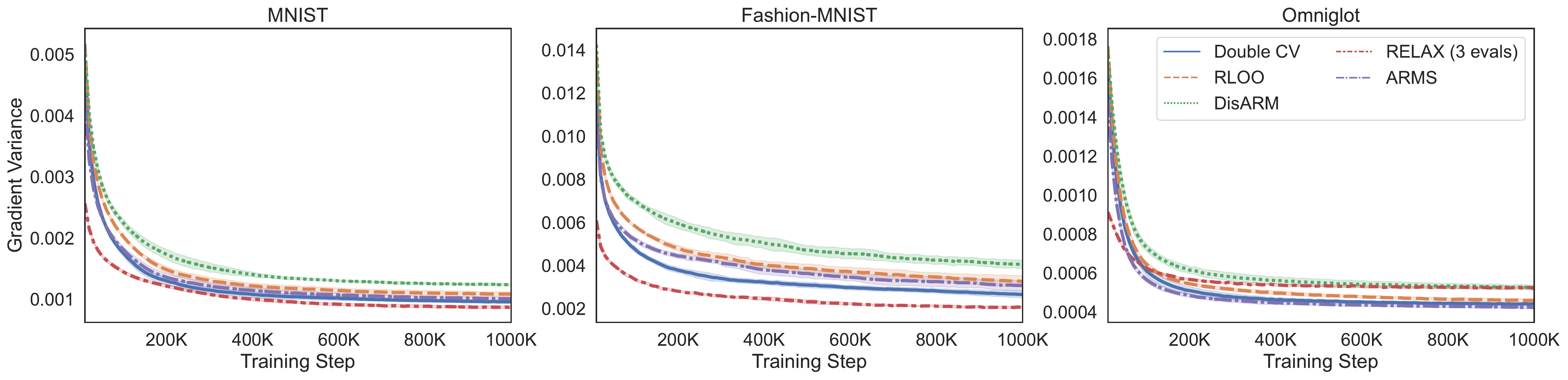}} \\
{\includegraphics[width=0.96\textwidth]
{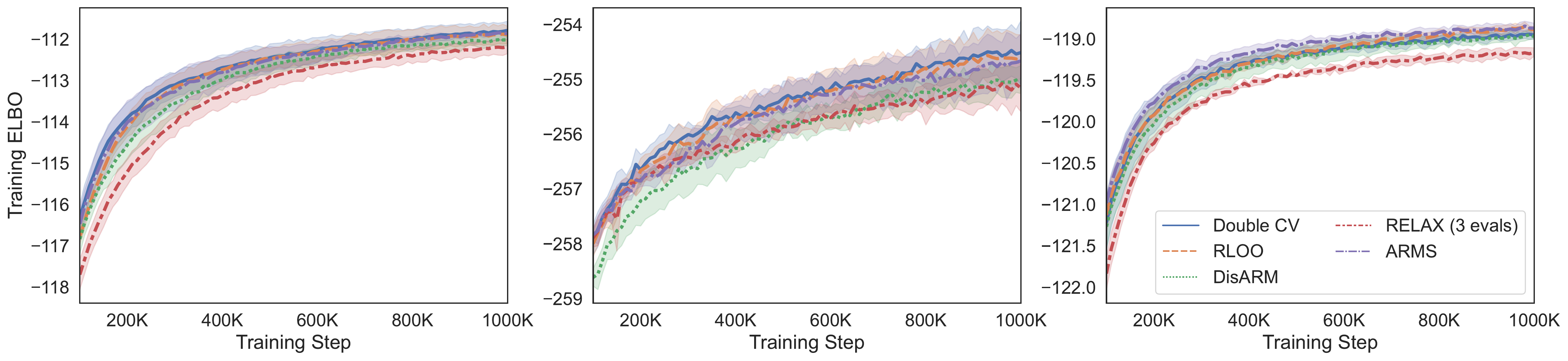}}
\end{tabular}
\caption{
Training linear binary latent VAEs with Bernoulli likelihoods with $K=4$ (except RELAX which needs 3 evaluations of $f$)
on dynamically binarized MNIST, Fashion-MNIST, and Omniglot.
\emph{Top:} Variance of gradient estimates.
 \emph{Bottom:} Average ELBO on training examples.
} 
\label{fig:dyn-linear-K4}
\end{figure*}

\begin{figure*}[ht]
\centering
\begin{tabular}{c}
{\includegraphics[width=0.96\textwidth]
{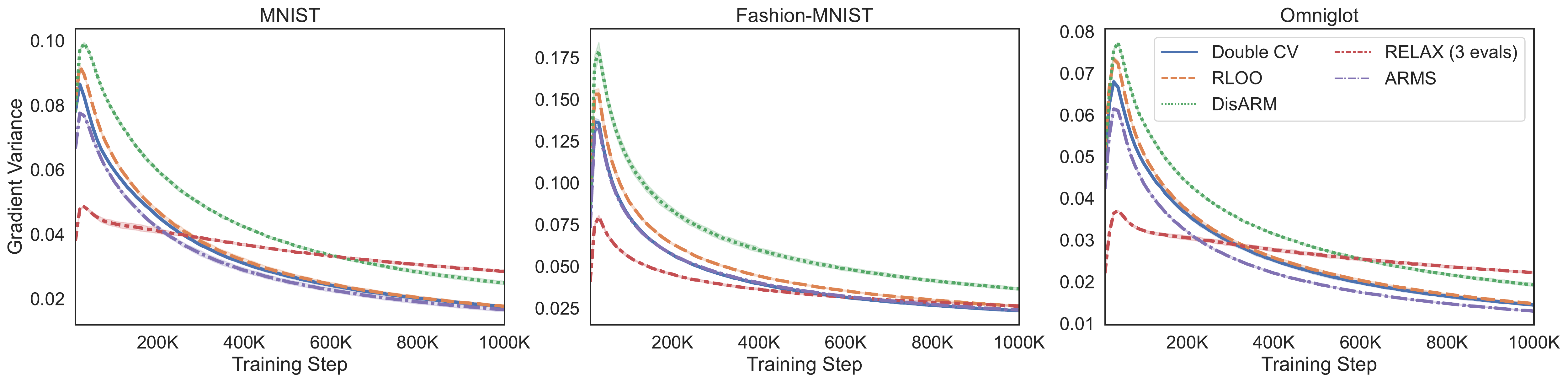}} \\
{\includegraphics[width=0.96\textwidth]
{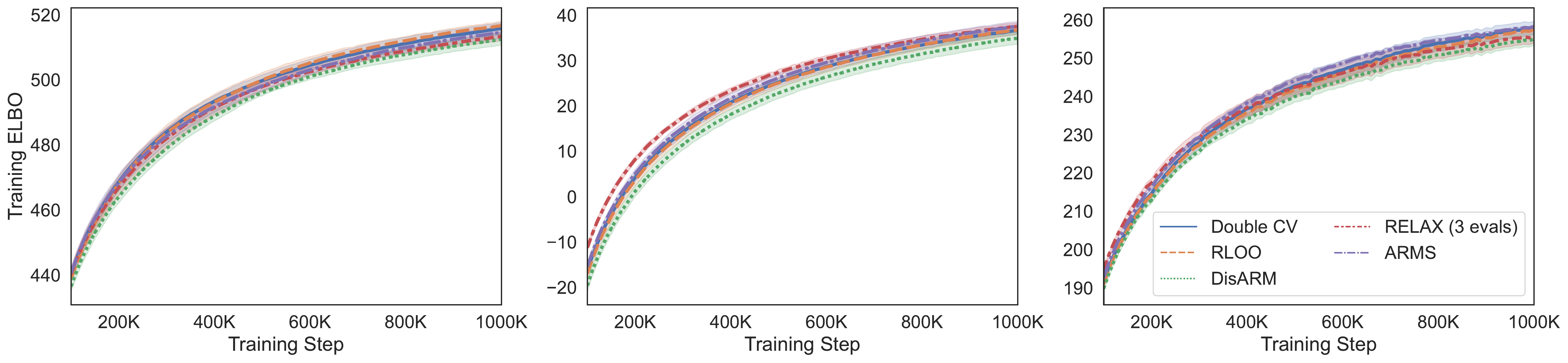}}
\end{tabular}
\caption{
Training linear binary latent VAEs with Gaussian likelihoods with $K=4$ (except RELAX which needs 3 evaluations of $f$) 
on non-binarized MNIST, Fashion-MNIST, and Omniglot.
\emph{Top:} Variance of gradient estimates.
 \emph{Bottom:} Average ELBO on training examples.
} 
\label{fig:cont-linear-K4}
\end{figure*}

We plot the gradient variance and average training ELBOs of training linear VAEs in Figures~\ref{fig:dyn-linear-K2},\ref{fig:cont-linear-K2},\ref{fig:dyn-linear-K4}, and \ref{fig:cont-linear-K4}.

\end{document}